%% file: main.tex
\documentclass[letter, 10 pt, conference]{ieeeconf}  

\IEEEoverridecommandlockouts 
\usepackage[english]{babel}
\usepackage[utf8]{inputenc}
\usepackage{times} 
\usepackage{cite}

\DeclareMathAlphabet{\pazocal}{OMS}{zplm}{m}{n}

\usepackage{hyperref}
\usepackage[nolist]{acronym}

\usepackage{multirow}

\usepackage{siunitx}
\usepackage{graphicx} 
\usepackage{epsfig} 
\usepackage{pgfplots}
\usepackage{caption}
\usepackage[font=footnotesize]{subcaption}
\usepackage{pgfplots}

\usepackage{float}

\usepackage[export]{adjustbox}

\usepackage{xcolor}
\definecolor{myYellow}{rgb}{0.93,0.69,0.13}
\definecolor{myPurple}{rgb}{0.49,0.18,0.56}
\definecolor{myGreen}{rgb}{0.26 0.72 0.54}

\usepackage{mathtools}
\usepackage{nicefrac}
\usepackage{amsmath}
\usepackage{amssymb}
\usepackage{bm} 

\DeclareMathOperator*{\minimize}{minimize}
\DeclareMathOperator*{\maximize}{maximize}

\usepackage{etoolbox}
\makeatletter%
\AfterPreamble{%
	\usepackage{hyperref}%
	\let\oldhypertarget\hypertarget%
	\renewcommand{\hypertarget}[2]{%
		\oldhypertarget{#1}{#2}%
		\protected@write\@mainaux{}{%
			\string\expandafter\string\gdef%
			\string\csname\string\detokenize{#1}\string\endcsname{#2}%
		}%
	}%
	\newcommand{\myhyperlink}[1]{%
		\hyperlink{#1}{\csname #1\endcsname}%
	}%
}
\makeatother%

\newcounter{Remark}
\newcounter{Definition}
\newcommand{\displayDefinitions}[2][]{%
	\stepcounter{Definition}%
	\textit{Definition~}\hypertarget{#1}{\theDefinition}\textit{~(#2)}%
}

\newcommand{\refDefinitions}[1][]{%
	Def.~\myhyperlink{#1}%
}

\newcounter{Problem}
\usepackage{algorithm}
\usepackage[noend]{algpseudocode}

\makeatletter
\def\BState{\State\hskip-\ALG@thistlm}
\makeatother

\usepackage{tikz}
\pgfdeclarelayer{foreground}
\pgfsetlayers{background, main, foreground}
\usetikzlibrary{quotes, angles, backgrounds, arrows, automata, shapes, positioning, calc, through, spy, decorations.pathreplacing, decorations.markings, arrows.meta, automata, petri}

\tikzset{
    imglabel/.style={
      rectangle,
      inner sep=2pt,
      text=black,
      minimum height=1em,
      text centered,
      fill=white,
      fill opacity=1.0,
      text opacity=1,
      anchor=south west,
    },
  }
\tikzset{
	state/.style={
		rectangle,
		draw=black, very thick,
		minimum height=1.0em,
		text centered,
	},
}
\tikzset{
  on each segment/.style={
    decorate,
    decoration={
      show path construction,
      moveto code={},
      lineto code={
        \path [#1]
        (\tikzinputsegmentfirst) -- (\tikzinputsegmentlast);
      },
      curveto code={
        \path [#1] (\tikzinputsegmentfirst)
        .. controls
        (\tikzinputsegmentsupporta) and (\tikzinputsegmentsupportb)
        ..
        (\tikzinputsegmentlast);
      },
      closepath code={
        \path [#1]
        (\tikzinputsegmentfirst) -- (\tikzinputsegmentlast);
      },
    },
  },
  mid arrow/.style={postaction={decorate,decoration={
        markings,
        mark=at position .5 with {\arrow[#1]{stealth}}
      }}},
}

\graphicspath{{./figures/}}

\newcommand\copyrighttext{%
    \small \begin{center} \color{red} \textcopyright\,2023 IEEE. Personal use of this material is permitted. Permission from IEEE must be obtained for all other uses, in any current or future media, including reprinting/republishing this material for advertising or promotional purposes, creating new collective works, for resale or redistribution to servers or lists, or reuse of any copyrighted component of this work in other works. \end{center}}
\newcommand\copyrightnotice{%
	\begin{tikzpicture}[remember picture,overlay]
	\node[anchor=south,yshift=25.6cm] at (current page.south) 
	{\color{red}\fbox{\parbox{\dimexpr\textwidth-\fboxsep-\fboxrule\relax}{\copyrighttext}}};
	\end{tikzpicture}%
}

\title{\copyrightnotice \LARGE \bf A Signal Temporal Logic Planner for Ergonomic\\Human–Robot Collaboration} 

\author{Giuseppe Silano$^1$, Amr Afifi$^2$, Martin Saska$^1$, and Antonio Franchi$^{2,3,4}$   
    \thanks{$^1$Faculty of Electrical Engineering, Department of Cybernetics, Czech Technical University in Prague, 12135 Prague, Czech Republic, (emails: {\tt\footnotesize \{giuseppe.silano, martin.saska\}@fel.cvut.cz}).} 
    \thanks{$^2$Robotics and Mechatronics Department, Electrical Engineering,  Mathematics, and Computer Science (EEMCS) Faculty, University of Twente, 7500 AE Enschede, The Netherlands, (emails: {\tt\small \{a.n.m.g.afifi, a.franchi\}@utwente.nl}).}
    \thanks{$^3$Department of Computer, Control and Management Engineering, Sapienza University of Rome, 00185 Rome, Italy, {\tt\footnotesize antonio.franchi@uniroma1.it}}
    \thanks{$^4$LAAS-CNRS, Universit\'{e} de Toulouse, 31000 Toulouse, France, {\tt\footnotesize antonio.franchi@laas.fr}}
    \thanks{This work was partially funded by the European Union's Horizon 2020 research and innovation programme AERIAL-CORE under grant agreement no. 871479, by the CTU grant no. SGS23/177/OHK3/3T/13, by the Czech Science Foundation (GAČR) within research project no. 23-07517S, and by the OP VVV funded project CZ.02.1.01/0.0/0.0/16 019/0000765 ``Research Center for Informatics".}
}

\begin{document}

\maketitle
\thispagestyle{empty} 
\pagestyle{empty} 


\begin{acronym}
    \acro{AGM}[AGM]{Arithmetic-Geometric Mean}
    \acro{AR}[AR]{Aerial Robot}
    \acro{CoM}[CoM]{Center of Mass}
    \acro{DoF}[DoF]{Degree of Freedom}
    \acro{GTMR}[GTMR]{Generically Tilted Multi-Rotor}
    \acro{HRI}[HRI]{Human-Robot Interaction}
    \acro{ILP}[ILP]{Integer Linear Programming}
    \acro{LSE}[LSE]{Log-Sum-Exponential}
    \acro{MILP}[MILP]{Mixed-Integer Linear Programming}
    \acro{MRAV}[MRAV]{Multi-Rotor Aerial Vehicle}
    \acro{NLP}[NLP]{Nonlinear Programming}
    \acro{NMPC}[NMPC]{Nonlinear Model Predictive Control}
    \acro{pHRI}[pHRI]{physical Human Robot Interaction}
    \acro{ROS}[ROS]{Robot Operating System}
    \acro{STL}[STL]{Signal Temporal Logic}
    \acro{TL}[TL]{Temporal Logic}
    \acro{UAV}[UAV]{Unmanned Aerial Vehicle}
    \acro{UAM}[UAM]{Unmanned Aerial Manipulator}
    \acro{VRP}[VRP]{Vehicle Routing Problem}
    \acro{wrt}[w.r.t.]{with respect to}
\end{acronym}



\begin{abstract}

This paper proposes a method for designing human-robot collaboration tasks and generating corresponding trajectories. The method uses high-level specifications, expressed as a~\ac{STL} formula, to automatically synthesize task assignments and trajectories. To illustrate the approach, we focus on a specific task: a multi-rotor aerial vehicle performing object handovers in a power line setting. The motion planner considers limitations, such as payload capacity and recharging constraints, while ensuring that the trajectories are feasible. Additionally, the method enables users to specify robot behaviors that take into account human comfort (e.g., ergonomics, preferences) while using high-level goals and constraints. The approach is validated through numerical analyzes in MATLAB and realistic Gazebo simulations using a mock-up scenario.

\end{abstract}






\section{Introduction}
\label{sec:introduction}

In the field of robotics,~\acp{AR} and~\acp{MRAV} have gained significant attention in recent 
years due to their impressive agility, maneuverability, and ability to be equipped with a 
variety of onboard sensors~\cite{OlleroTRO2022}. Their modular design and versatility have 
made them suitable for a wide range of applications, including those which involve either 
contactless~\cite{Alcantara2020IEEEAccess} or physical interaction with their 
surroundings~\cite{Tognon2019RAL}.

There are many real-world examples where the use of aerial robots is advantageous, such as for work environments at heights, wind turbines, large construction sites, or power transmission lines~\cite{SilanoICUAS2021, BonillaICUAS23}. These types of settings often require specialized and trained personnel to use expensive equipment and special vehicles. The use of aerial robots as robotic co-workers~\cite{Afifi2022ICRA, Corsini2022IROS} in these scenarios can, for example, facilitate tasks by flying to the target location and carrying tools, reducing the physical and cognitive load on human operators, etc.. However, in order to realize these benefits, it is important to consider human ergonomics and safety~\cite{WojciechowskaHRI2019} when designing aerial robots, particularly multi-rotors.

Despite the potential of~\acp{MRAV} to work closely with human operators, their use in such scenarios is still limited. In contrast, there is a wealth of research on~\ac{HRI} involving ground robots and human partners~\cite{Ajoudani2018AR}. For example, previous studies have looked at the use of manipulators to assist humans in handling heavy or bulky objects, or during assembly tasks~\cite{Gienger2018IROS}. The issue of object handover has also been extensively studied in the literature~\cite{Ortenzi2021TRO}.

Therefore, to allow~\acp{AR} to effectively collaborate with human workers and address critical ergonomic and safety concerns while additionally minimizing the physical and cognitive demands on human operators, advanced task and motion planning techniques are required.~\ac{TL} can fulfill this role by providing a mathematical framework for expressing complex specifications that combine natural language commands with temporal and Boolean operators~\cite{Belta2007RAM}. In particular,~\acf{STL}~\cite{maler2004FTMATFTS} is endowed with a metric called \textit{robustness}, which not only allows one to determine whether a system's execution satisfies certain requirements, but also provides a measure of how well these requirements have been satisfied. This leads to an optimization problem aimed at maximizing the robustness score, thereby generating the optimal feasible trajectory for the system that meets desired specifications.

\begin{figure}[tb]
    \centering
    \adjincludegraphics[width=0.65\columnwidth, trim={{0.22\width} {0.20\height} {0.0\width} {.0\height}}, clip]{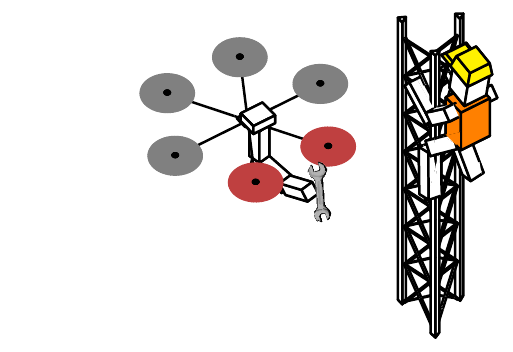}
    \vspace{-0.6em}
    \caption{A schematic representation of an~\ac{MRAV} delivering a tool to a human worker in a power line scenario.}
    \label{fig:handover_representation}
\end{figure}

In this paper, we present an~\ac{MRAV} motion planner that leverages~\ac{STL} specifications to facilitate ergonomic human-robot collaboration. As a motivating example, we consider the task of an~\ac{MRAV} performing object handovers in a power line scenario, as depicted in Figure~\ref{fig:handover_representation}. The mission requirements are expressed as an~\ac{STL} formula. We maximize its robustness by formulating a nonlinear non-convex max-min optimization problem. To address the complexity of this nonlinear optimization, we employ a hierarchical approach that first solves an~\ac{ILP} problem, and then passes the result to the final~\ac{STL} optimizer.

\subsection{Related work}
\label{sec:relatedWork}

The process of handover typically encompasses multiple stages, including the \textit{approach}, \textit{reach}, and \textit{transfer} phases, as noted in previous studies~\cite{Corsini2022IROS, Afifi2022ICRA}. Many researchers have typically studied each phase individually, but there are a few notable exceptions. In~\cite{Medina2016Humanoids}, a control architecture for fluid handovers that addresses all phases in a cohesive manner is proposed. This approach takes into account the interactions that occur during the handover and specifically aims to minimize unwanted wrench components that are not essential for moving and holding the object. However, the proposed control scheme does not explicitly factor in safety and ergonomics, which are crucial considerations in a framework designed for collaboration between aerial robots and humans, particularly in high-risk environments. 

The inclusion of human comfort and ergonomics into robot control and planning software has also been explored in the literature. One of the earliest works in this area is~\cite{Sisbot2012TRO}, which develops a manipulation planner accounting for various human factors, such as ergonomics and field of view. Subsequent studies have advanced this approach. For example,~\cite{Peternel2017Humanoids} proposes a method for computing human joint torques based on a whole-body dynamic model and then controlling a ground mobile manipulator to minimize the overloading of human joints. However, none of these works consider~\acp{AR} as robotic co-workers interacting with a human operator.

In addition to the control and planning aspects, some studies concentrate on improving these elements by equipping~\acp{AR} with the ability to perceive the human subject through sensors. This is crucial as any reduction in visibility of the human collaborator could negatively impact the task. Perception-constrained control is a key consideration in these studies. For example,~\cite{Corsini2022IROS} proposes an~\ac{NMPC} formulation that incorporates human ergonomics and comfort as objectives, while also enforcing perception and actuation limits. 
Other research focuses on using dynamic programming to ensure safety so that the robot can never cause injury to the interacting human. For instance,~\cite{Afifi2022ICRA} presents a comprehensive framework formulated as a constrained quadratic programming problem for controlling an aerial manipulator during physical interactions with a human operator. 
While these approaches enable~\acp{AR} to interact safely and ergonomically with a single human operator, they do not account for scenarios where multiple operators are involved and tools need to be carried multiple times throughout the mission. 

Alternatively, the field of automatic synthesis of robot controllers for human-robot handovers from formal specifications has also been explored. For example, the authors in~\cite{Kshirsagar2019IROS} present a controller for human-robot handovers automatically generated from high-level specifications in~\ac{STL}. This approach offers formal guarantees on the timing of each handover phase. 
In the realm of formal methods,~\cite{Webster2019ArXiv} uses probabilistic model-checking in a human-robot handover task, validating a controller~\ac{wrt} safety and liveness~\cite{Alpern1985IPL} specifications. Finally, the authors in~\cite{Kshirsagar2019IROS} describe a formalism for human-in-the-loop control synthesis and set up a semi-autonomous controller from~\ac{TL} specifications. However, as far as the authors are aware, this is the first study that addresses the task assignment and trajectory generation problem for~\acp{AR}, specifically for~\acp{MRAV}, to enhance human-robot ergonomic collaboration.



\subsection{Contributions}
\label{sec:contributions}

This paper presents a novel method in designing tasks and motion planning for human-robot collaboration prioritizing ergonomics. The approach is demonstrated using the specific task of an~\ac{MRAV} performing object handovers in a power line setting. The method employs~\ac{STL} to generate optimal trajectories that comply with the vehicle's dynamics and velocity and acceleration limits, while also fulfilling various mission specifications, such as collision avoidance and safety requirements. A hierarchical strategy is implemented to address the complexity of the problem, where an initial guess solution is obtained by solving an~\ac{ILP} problem to act as the starting point for the~\ac{STL} optimizer. This strategy builds upon prior work~\cite{SilanoRAL2021, CaballeroAR2022} by allowing users to specify robot behaviors that take into account human comfort and preferences using high-level goals and constraints. This includes computing trajectories that consider payload capacity limitations and refilling stations for longer-duration operations. Additionally, a method for computing the initial solution for the optimization problem is proposed.

The proposed approach offers several key benefits: (i) by using~\ac{STL} formulae, the framework can take into account explicit time requirements, making it easy to adapt and customize for various applications; (ii) the concise and unambiguous~\ac{STL} formulation enables end-users to specify robot behavior in terms they can understand, such as the direction of approach and zones to avoid, reducing the need for hand-coded algorithms; (iii) using automatic synthesis based on formal models, our approach provides timing guarantees. This means that the vehicle will always obey timing constraints, as long as the human behavior allows for it. It also notifies users of any constraint violations.
    
    
The approach has been validated through numerical simulations in MATLAB. Additionally, Gazebo simulations were used to demonstrate the approach's effectiveness in a scenario that closely resembles real-world implementation.




\section{Problem Description}
\label{sec:problemDescription}

\begin{figure}[tb]
    \centering
    \adjincludegraphics[trim={{.08\width} {.07\height} {.08\width} {.08\height}}, clip, width=\columnwidth ]{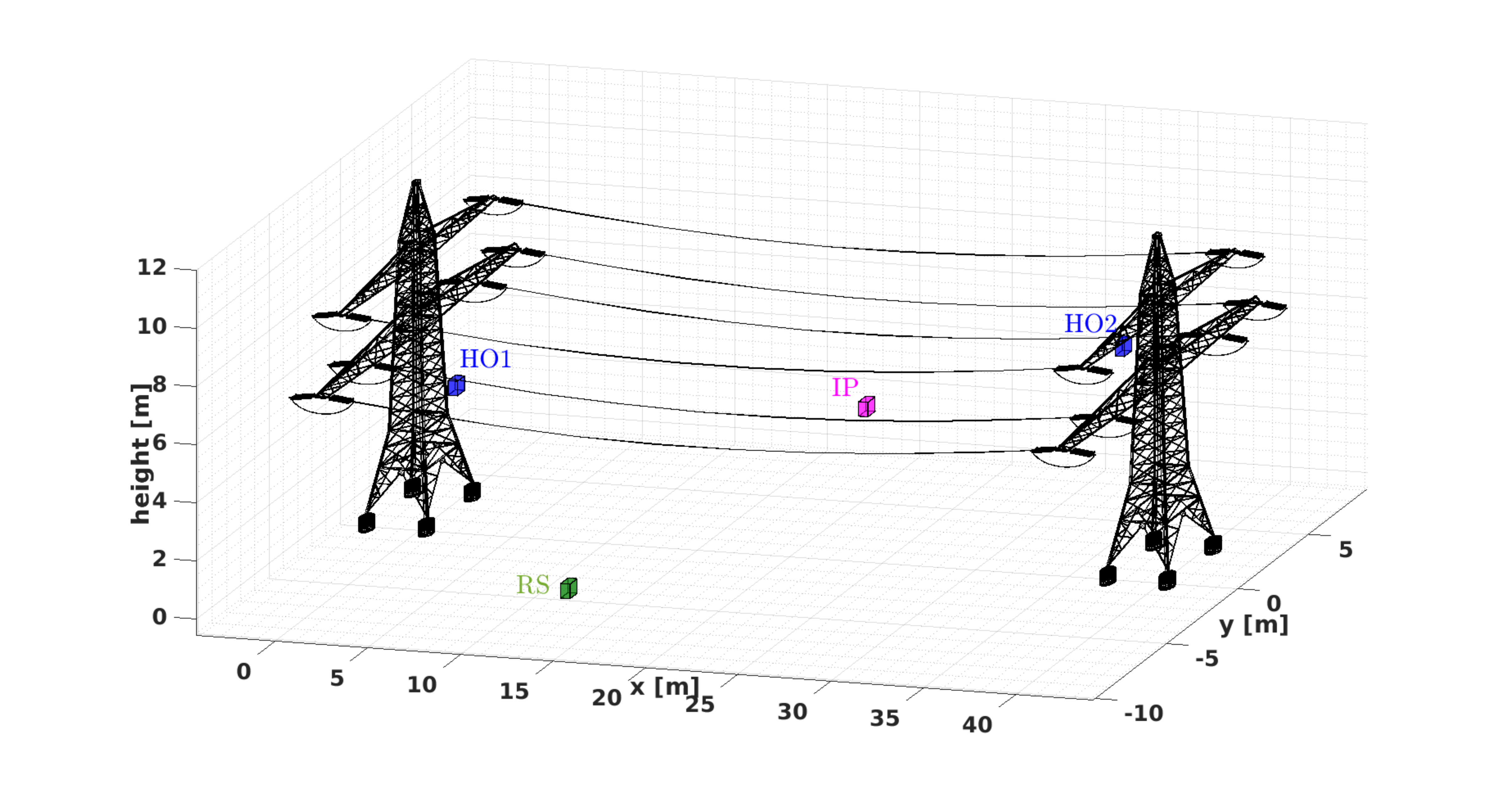}
    \caption{Power line scenario for the ergonomic human-robot collaboration. Human operators (HO) are represented in blue, while the~\ac{MRAV}’s initial position (IP) and the refilling station (RS) are depicted in magenta and green, respectively.}
    \label{fig:scenarioSTL}
\end{figure}

The focus of this paper is to enhance the ergonomic collaboration between humans and robots. The scenario under examination is that of an~\ac{MRAV} equipped with a manipulation arm performing object handovers in a power line setting. This scenario is depicted in Figure~\ref{fig:scenarioSTL}. The objective is to design a trajectory for the drone that takes into account human ergonomic needs by approaching the operator from the front, either from the left or right, from above or below, and never from behind~\cite{WojciechowskaHRI2019}. To simplify the problem, the location of the handover operation is represented as a 3D space for each operator. We assume that the~\ac{MRAV} begins the mission with a tool. For ease of understanding, we consider that only one object can be delivered at a time. However, the method can be easily extended to multiple objects. Once the~\ac{MRAV} reaches the operator, it is assumed that an onboard low-level controller handles the handover procedure, e.g.~\cite{Afifi2022ICRA, Corsini2022IROS}. The~\ac{MRAV} has limited velocity, acceleration, and payload capacity, meaning the number of tools it can carry is restricted. Furthermore, there is a refilling station on the ground along the power line where the drone can reload tools and resume operation. The goal is to plan a trajectory for the~\ac{MRAV} to complete the mission within a specified maximum time frame, while also satisfying dynamics and capacity constraints. Further, safety requirements must be met, such as avoiding obstacles and never approaching the operator from behind. It is assumed that a map of the environment, including a polyhedral representation of obstacles like power towers and cables, is known in advance.

\section{Preliminaries}
\label{sec:preliminaries}


Let us consider a discrete-time dynamical system $\mathcal{S}$ represented in the form $x_{k+1}=f(x_k, u_k)$, where $x_{k+1}$, $x_k \in \mathcal{X} \subset \mathbb{R}^n$ are the next and current states of the system $\mathcal{S}$, respectively, and $u_k \in \mathcal{U} \subset \mathbb{R}^m$ is the control input. Let us also assume that $f \colon \mathcal{X} \times \mathcal{U} \rightarrow \mathcal{X}$ is differentiable in both arguments and locally Lipschitz. Therefore, given an initial state $x_0 \in \mathcal{X}_0 \subset \mathbb{R}^n$ and a time vector $\mathbf{t} = (t_0, \dots, t_N)^\top \in \mathbb{R}^{N+1}$, with $N \in \mathbb{N}_{>0}$ being the number of samples that describe the evolution of the system $\mathcal{S}$ with sampling period $T_s \in \mathbb{R}_{>0}$, we can define the finite control input sequence $\mathbf{u}=(u_0, \dots, u_{N-1})^\top \in \mathbb{R}^N$ as the input to provide to the system $\mathcal{S}$ to attain the unique sequence of states $\mathbf{x}=(x_0, \dots, x_N)^\top \in \mathbb{R}^{N+1}$.


Let us consider a~\ac{GTMR} model~\cite{MichielettoTRO2018} to describe the vehicle's dynamics. Also, let us denote with $\mathcal{F}_W$ and $\mathcal{F}_B$ the world frame and body frame reference systems, respectively. The body frame is attached to the~\ac{GTMR} such that the origin of the frame $\mathcal{O}_B$ coincides with the~\ac{CoM} of the vehicle. The position of the origin $\mathcal{O}_B$ of the body frame $\mathcal{F}_B$~\ac{wrt} the world frame $\mathcal{F}_W$ is denoted with $\mathbf{p} = (p^{(1)}, p^{(2)}, p^{(3)})^\top \in \mathbb{R}^3$, while the velocity and acceleration of $\mathcal{O}_B$ in $\mathcal{F}_W$ are denoted with $\mathbf{v} = (v^{(1)}, v^{(2)}, v^{(3)})^\top \in \mathbb{R}^3$ and $\mathbf{a}  = (a^{(1)}, a^{(2)}, a^{(3)})^\top \in \mathbb{R}^3$, respectively.

Hence, we can define the state sequence $\mathbf{x}$ and the control input sequence $\mathbf{u}$ of a~\ac{GTMR} model as $\mathbf{x}=(\mathbf{p}^{(1)}, \mathbf{v}^{(1)}, \mathbf{p}^{(2)}, \mathbf{v}^{(2)}, \mathbf{p}^{(3)}, \mathbf{v}^{(3)})^\top$ and $\mathbf{u}=(\mathbf{a}^{(1)}, \mathbf{a}^{(2)}, \mathbf{a}^{(3)})^\top$, where $\mathbf{p}^{(j)}$, $\mathbf{v}^{(j)}$, and $\mathbf{a}^{(j)}$, with $j=\{1, 2, 3\}$, represent the sequences of position, velocity, and acceleration of the vehicle along the $j$-axis of the world frame $\mathcal{F}_W$, respectively. 

Finally, let us denote with $p_k^{(j)}$, $v_k^{(j)}$, $a_k^{(j)}$, and $t_k$, with $k \in \mathbb{N}_{\geq 0}$, the $k$-th element of the sequences $\mathbf{p}^{(j)}$, $\mathbf{v}^{(j)}$, $\mathbf{a}^{(j)}$, and vector $\mathbf{t}$, respectively.



\subsection{Signal temporal logic}
\label{sec:signalTemporalLogic}

\displayDefinitions[SignalTemporalLogic]{\acl{STL}}:~\ac{STL} is a concise and unambiguous specification language for describing the temporal behavior of real-valued signals~\cite{maler2004FTMATFTS}. 
One major benefit of using~\ac{STL} formulae for motion planning over traditional algorithms~\cite{LaValleBook} is that all mission specifications can be encapsulated into a single formula $\varphi$. For example, the statement ``at least two vehicles must complete tasks $A$ and $B$ before task $C$ is performed, one of them must execute task $D$ within the time interval $[t_1, t_2]$, while all of them must avoid obstacles and remain within the designated workspace" can be expressed into a single~\ac{STL} formula $\varphi$. The syntax and semantics of~\ac{STL} are detailed in~\cite{maler2004FTMATFTS, donze2010ICFMATS}, however a brief overview is provided here.~\ac{STL}'s grammar enables the representation of complex behavioral requirements using temporal operators, such as \textit{until} ($\mathcal{U}$), \textit{always} ($\square$), and \textit{eventually} ($\lozenge$), as well as logical operators like \textit{and} ($\wedge$), \textit{or} ($\vee$), and \textit{negation} ($\neg$). These operators act on atomic propositions (also known as \textit{predicates}), which are simple statements or assertions that are either \textit{true} ($\top$) or \textit{false} ($\bot$). Examples of atomic propositions include belonging to a particular region or comparing real values (e.g., a distance exceeding a threshold). An~\ac{STL} formula $\varphi$ is considered valid if it evaluates to a \textit{true} ($\top$) logic value, and invalid ($\bot$) otherwise. For instance, informally, the formula $\varphi_1 \mathcal{U}_I \varphi_2$ means that $\varphi_2$ must hold at some point within the time interval $I$ and, until then, $\varphi_1$ must hold without interruption.



\subsection{Robust signal temporal logic}
\label{sec:robustSignalTemporalLogic}

\displayDefinitions[RobustSignalTemporalLogic]{\ac{STL} Robustness}: Uncertainties, dynamic conditions, and unexpected events can all impact the satisfaction of an~\ac{STL} formula $\varphi$ (\refDefinitions[SignalTemporalLogic]). To account for these factors and ensure a margin of satisfaction for an~\ac{STL} formula $\varphi$, the concept of \textit{robust semantics} for~\ac{STL} formulae has been developed~\cite{maler2004FTMATFTS, donze2010ICFMATS, Fainekos2009TCS}. This \textit{robustness}, represented by $\rho$, is a quantitative metric that helps guide the optimization process towards finding the best feasible solution for meeting the statement (mission) requirements. It can be formally defined using the following recursive formulae:
\begin{equation*}
    \begin{array}{rll}
    \rho_{p_i} (\mathbf{x}, t_k) & = & \mu_i (\mathbf{x}, t_k), \\ 
    \rho_{\neg \varphi} (\mathbf{x}, t_k) & = & - \rho_\varphi (\mathbf{x}, t_k), 
    \\
    \rho_{\varphi_1 \wedge \varphi_2} (\mathbf{x}, t_k) & = & \min \left(\rho_{\varphi_1} (\mathbf{x}, t_k), \rho_{\varphi_2} (\mathbf{x}, t_k) \right), \\
    \rho_{\varphi_1 \vee \varphi_2} (\mathbf{x}, t_k) & = & \max \left(\rho_{\varphi_1} (\mathbf{x}, t_k), \rho_{\varphi_2} (\mathbf{x}, t_k) \right), \\
    \rho_{\square_I \varphi} (\mathbf{x}, t_k) & = & \min\limits_{t_k^\prime \in [t_k + I]} \rho_\varphi (\mathbf{x}, t_k^\prime), \\
    \rho_{\lozenge_I \varphi} (\mathbf{x}, t_k) & = & \max\limits_{t_k^\prime \in [t_k + I]} \rho_\varphi (\mathbf{x}, t_k^\prime), \\
    \rho_{\varphi_1 \mathcal{U}_I \varphi_2} (\mathbf{x}, t_k) & = & \max\limits_{t_k^\prime \in [t_k + I]} \Bigl( \min \left( \rho_{\varphi_2} (\mathbf{x}, t_k^\prime) \right), \\
    & &  \hfill \min\limits_{ t_k^{\prime\prime} \in [t_k, t_k^\prime] } \left( \rho_{\varphi_1} (\mathbf{x}, t_k^{\prime \prime} \right)  \Bigr),
    \end{array}
\end{equation*}
where $t_k + I$ represents the Minkowski sum of the scalar $t_k$ and the time interval $I$. The formulae above consist of a set of \textit{predicates}, $p_i$, along with their corresponding real-valued function $\mu_i(\mathbf{x}, t_k)$, each of which is considered true if its robustness value is greater than or equal to zero, and false otherwise. The whole formula behaves like a logical formula, which is deemed false if any of the predicates are false. As an illustration, using the previous example of belonging to a particular region, we can first divide the mission into a set of predicates, i.e., bounds on distance relative to upper and lower margins. Next, we compute the robustness value associated with each predicate (how well or poorly the specification is being satisfied). Finally, all predicates are evaluated using the logical formulae presented earlier. The result is a numerical value that indicates whether the specification is being met and to what degree. Further information on this can be found in~\cite{maler2004FTMATFTS, donze2010ICFMATS, Fainekos2009TCS}. In this case, we say that $\mathbf{x}$ satisfies the~\ac{STL} formula $\varphi$ at time $t_k$ if $\rho_\varphi(\mathbf{x}, t_k) > 0$, and $\mathbf{x}$ violates $\varphi$ if $\rho_\varphi(\mathbf{x}, t_k) \leq 0$.

Therefore, we can compute the control inputs $\mathbf{u}$ that maximize robustness over the set of finite state and input sequences $\mathbf{x}$ and $\mathbf{u}$, respectively. This optimal sequence $\mathbf{u}^\star$ is considered valid if $\rho_\varphi (\mathbf{x}^\star, t_k)$ is positive, with $\mathbf{x}^\star$ and $\mathbf{u}^\star$ satisfying the dynamics of the system. The larger the value of $\rho_\varphi (\mathbf{x}^\star, t_k)$, the more robust the system's behavior is considered to be.

\displayDefinitions[SmoothApproximation]{Smooth Approximation}: A limitation of the standard robustness measure is that it is non-smooth and non-convex for many useful specifications due to the inclusion of $\min$ and $\max$ operators. To overcome this issue, recent research has focused on smooth approximations of the robustness measure $\tilde{\rho}_\varphi(\mathbf{x}, t_k)$~\cite{Pant2017CCTA, MehdipourACC2019}. These approximations allow for the use of efficient gradient-based optimization methods and have been shown to perform well on a wide range of problems, providing significant speed and scalability improvements~\cite{LindemannAutomatica2019}. 



One of the approximations of the robustness measures is the~\ac{AGM} robustness. This approach retains many of the computational benefits of the most commonly used~\ac{LSE} method~\cite{Pant2017CCTA}. 
Furthermore,~\ac{AGM} robustness is more conservative in the sense that trajectories derived using this approach tend to be more robust to external disturbances. However, the~\ac{AGM} robustness is almost smooth everywhere, i.e., its analytical gradient does not always exist. 
While this issue with singularities may pose challenges for the solver, it is still important to keep in mind that the optimization should not reach the boundary conditions. For example, in the case of belonging to the workspace area, there is no reason for the vehicle to be at the boundaries. In light of this, we choose the~\ac{AGM} as smooth approximation for the robustness measure $\tilde{\rho}_\varphi (\mathbf{x}, t_k)$. The full description of the~\ac{AGM} robustness syntax and semantics can be found in~\cite{MehdipourACC2019} and is not given here for the sake of brevity. 

\displayDefinitions[STL-Planner]{\ac{STL} Motion Planner}: By encoding the mission specifications detailed in Section~\ref{sec:problemDescription} as an~\ac{STL} formula $\varphi$ and replacing its robustness $\rho_\varphi(\mathbf{x}, t_k)$ with the smooth approximation $\tilde{\rho}_\varphi(\mathbf{x}, t_k)$ (as defined in~\refDefinitions[SmoothApproximation]), the problem of generating trajectories for the~\ac{GTMR} model can be formulated as the optimization problem~\cite{SilanoRAL2021}:
\begin{equation}\label{eq:optimizationProblemMotionPrimitives}
    \begin{split}
    &\maximize_{\mathbf{p}^{(j)}, \mathbf{v}^{(j)},\,\mathbf{a}^{(j)}} \;\;
    {\tilde{\rho}_\varphi (\mathbf{p}^{(j)}, \mathbf{v}^{(j)} )} \\
    &\quad \,\;\, \text{s.t.}~\quad\;\;\; \lvert v^{(j)}_k \rvert \leq 
    \bar{v}^{(j)}, \lvert a^{(j)}_k \vert  \leq \bar{a}^{(j)}, \\
    &\,\;\;\;\;\, \qquad \quad\;\;\, \mathbf{S}^{(j)}, \forall k=\{0,1, \dots, N-1\}
    \end{split},
\end{equation}
where $\bar{v}^{(j)}$ and $\bar{a}^{(j)}$ are the maximum desired values of velocity and acceleration along the motion, respectively, and $\mathbf{S}^{(j)} ( p_k^{(j)}, v_k^{(j)}, a_k^{(j)} ) = (p_{k+1}^{(j)}, v_{k+1}^{(j)}, a_{k+1}^{(j)})^\top$ are the vehicle motion primitives along each $j$-axis of the world frame $\mathcal{F}_W$ encoding the splines presented in~\cite[eq.(2)]{SilanoRAL2021}. 



\section{Problem Solution}
\label{sec:problemSolution}

In this section, we apply the~\ac{STL} framework presented in Section~\ref{sec:preliminaries} to formulate the optimization problem in Section~\ref{sec:problemDescription}, resulting in a nonlinear non-convex max-min problem formulated as a~\ac{NLP} problem solved through dynamic programming (Section~\ref{sec:motionPlanner}). 
To find a solution for this type of nonlinear problem in a reasonable amount of time, we generate an initial guess from a simplified~\ac{ILP} formulation (Section~\ref{sec:initialGuess}), that draws inspiration from prior research~\cite{CaballeroAR2022}. It is worth noting that the proposed approach distinguishes itself from previous efforts by incorporating ergonomics features while maintaining consistency with the previously identified objectives. 



\subsection{Motion planner}
\label{sec:motionPlanner}

\begin{figure}[tb]
    \centering
    \adjincludegraphics[width=0.65\columnwidth, trim={{0.22\width} {0.025\height} {0.0\width} {0\height}}, clip]{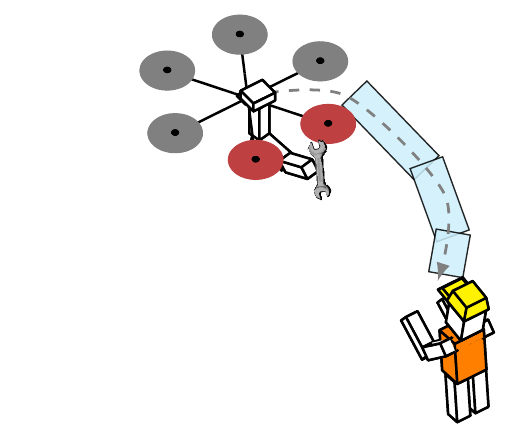}
    \vspace{-0.7em}
    \caption{An illustrative representation of an~\ac{MRAV} approaching a human operator. In blue are depicted the areas ($\underline{p}_\mathrm{pr}^{(j)}, \bar{p}_\mathrm{pr}^{(j)}$) where the drone can approach the operator, while in gray is a sample output from the~\ac{STL} optimizer.}
    \label{fig:scenarioHumanApproach}
\end{figure}

In this section, the requirements for the problem outlined in Section~\ref{sec:problemDescription} are translated into the~\ac{STL} formula, $\varphi$. The mission to perform object handovers with an~\ac{MRAV} have two types of specifications. Firstly, safety must be maintained throughout the entire mission duration, $t_N$; the~\ac{MRAV} must remain within the designated area ($\varphi_\mathrm{ws}$), avoid collisions with surrounding objects ($\varphi_\mathrm{obs}$), and never approach to the operator from behind ($\varphi_\mathrm{beh}$). Secondly, certain ergonomic-related objectives must be met at specific times in the mission duration; each human operator must be visited once by the~\ac{MRAV} and the vehicle must stay there for $t_\mathrm{han}$ ($\varphi_\mathrm{han}$), with $t_\mathrm{han} \ll t_N$, to perform the object handover. The~\ac{MRAV} must approach the operator from the front, either from the left or right, from above or below, based on the operator's preferences ($\varphi_\mathrm{pr}$). Additionally, due to the limited carrying capacity of the~\ac{MRAV}, the vehicle must stop at a refilling station and remain there for a duration of $t_\mathrm{rs}$
, with $t_\mathrm{rs} \ll t_N$, once its onboard supply of tools is depleted in order to replenish its supply ($\varphi_\mathrm{cap}$). Lastly, after completing handover operations, the~\ac{MRAV} must fly to its nearest refilling station ($\varphi_\mathrm{rs}$). All mission requirements can be expressed in the~\ac{STL} formula:
\begin{equation}\label{eq:stlFormula}
    \resizebox{0.91\hsize}{!}{$%
    \varphi = \square_{t_N} \varphi_\mathrm{ws} \wedge \varphi_\mathrm{obs} \wedge \varphi_\mathrm{beh} \bigwedge \lozenge_{t_N} \left( \varphi_\mathrm{han} \wedge \varphi_\mathrm{pr} \wedge \varphi_\mathrm{cap} \right) \pazocal{U}_{t_N} \varphi_\mathrm{rs},
    $}%
\end{equation}
with
\begin{subequations}\label{eq:STLcomponents}
    \begin{align}
    \varphi_\mathrm{ws} &= \mathbf{p}^{(j)} \in (\underline{p}^{(j)}_\mathrm{ws}, \bar{p}^{(j)}_\mathrm{ws}), \label{subeq:belongWorkspace} \\
    \varphi_\mathrm{obs} &= \bigwedge_n^{\mathrm{obs}} \, \mathbf{p}^{(j)} \not\in ({^n}\underline{p}_\mathrm{obs}^{(j)}, {^n}\bar{p}_\mathrm{obs}^{(j)}), \label{subeq:avoidObostacles} \\
    \varphi_\mathrm{beh} &= \bigwedge_n^{\mathrm{beh}} \, \mathbf{p}^{(j)} \not\in ({^n}\underline{p}_\mathrm{beh}^{(j)}, {^n}\bar{p}_\mathrm{beh}^{(j)}), \label{subeq:behind} \\
    \varphi_\mathrm{han} &= \bigwedge_n^{\mathrm{ho}} \, \square_{t_\mathrm{han}} \, \mathbf{p}^{(j)} \in ({^n}\underline{p}_\mathrm{ho}^{(j)}, {^n}\bar{p}_\mathrm{ho}^{(j)}), \label{subeq:handover} \\
    \varphi_\mathrm{pr} &= \bigwedge_n^{\mathrm{ho}}  \, \left( \bigwedge_m^{\mathrm{pr}} \, \lozenge_{t_N} \, \mathbf{p}^{(j)} \in ({^{n,m}}\underline{p}_\mathrm{pr}^{(j)}, {^{n,m}}\bar{p}_\mathrm{pr}^{(j)}) \right), \label{subeq:preferences} \\
    \varphi_\mathrm{cap} &= \square_{t_\mathrm{rs}}(c == 0) \, \mathbf{p}^{(j)} \in (\underline{p}_\mathrm{rs}^{(j)}, \bar{p}_\mathrm{rs}^{(j)}), \label{subeq:refill} \\
    \varphi_\mathrm{rs} & = \mathbf{p}^{(j)} \in (\underline{p}_\mathrm{rs}^{(j)}, \bar{p}_\mathrm{rs}^{(j)}), \label{subeq:backHome} 
    \end{align}
\end{subequations}
where equation~\eqref{subeq:belongWorkspace} ensures that the position of the~\ac{MRAV}, denoted by $\mathbf{p}^{(j)}$, remains within the designated workspace. The minimum and maximum values of the workspace along the $j$-axis of the world frame $\mathcal{F}_W$ are represented by $\underline{p}^{(j)}_\mathrm{ws}$ and $\bar{p}^{(j)}_\mathrm{ws}$, respectively. Equations~\eqref{subeq:avoidObostacles},~\eqref{subeq:behind},~\eqref{subeq:handover},~\eqref{subeq:refill}, and~\eqref{subeq:backHome} establish guidelines for obstacle avoidance, operator safety, handover operations, payload capacity, and mission completion, respectively. The payload capacity of the~\ac{MRAV} is represented by $c \in \mathbb{N}_{>0}$ as a positive integer. The vertices of the rectangular regions 
identifying obstacles, areas behind the operators, operators themselves, and refilling stations along the $j$-axis of the world frame $\mathcal{F}_W$ are represented by $\underline{p}_\mathrm{obs}^{(j)}$, $\underline{p}_\mathrm{beh}^{(j)}$, $\underline{p}_\mathrm{ho}^{(j)}$, $\underline{p}_\mathrm{rs}^{(j)}$, $\bar{p}_\mathrm{obs}^{(j)}$, $\bar{p}_\mathrm{beh}^{(j)}$, $\bar{p}_\mathrm{ho}^{(j)}$ and $\bar{p}_\mathrm{rs}^{(j)}$, respectively. Finally, equation~\eqref{subeq:preferences} accounts for the human operators' preferences for the drone's approach. These preferences include the operator's preferred approach direction, such as front, left, right, above, or below~\cite{WojciechowskaHRI2019}. The preferences define specific areas represented by rectangular regions ($\underline{p}_\mathrm{pr}^{(j)}, \bar{p}_\mathrm{pr}^{(j)}$) along the $j$-axis within the world frame $\mathcal{F}_W$ where the drone is permitted to approach the operator. These areas are established by taking the operator's orientation into account and reflecting the potential paths that the drone's trajectory can take when approaching the operator from different directions (i.e., front, left, right, above, or below). 
The use of the \textit{eventually} operator $\lozenge_{t_N}$ guarantees that the predicate~\eqref{subeq:preferences}, i.e., the drone position $\mathbf{p}^{(j)}$ belonging to those rectangular regions, will hold somewhere within the time interval $t_N$. A schematic representation of the scenario is reported in Figure~\ref{fig:scenarioHumanApproach}.

By using the specifications described in~\eqref{eq:stlFormula}, we formulated the optimization problem described in~\refDefinitions[STL-Planner] to determine feasible trajectories that maximize the smooth robustness $\tilde{\rho}_\varphi(\mathbf{x}, t_k)$~\ac{wrt} the given mission specifications $\varphi$. In order to accomplish this, we must compute the robustness score for each individual predicate. The predicates within the~\ac{STL} formula~\eqref{eq:stlFormula} assess whether the~\ac{MRAV}'s position falls within a specific region, as described in~\eqref{subeq:belongWorkspace}--\eqref{subeq:backHome}. When the~\ac{MRAV} is located within a region, it is assigned a positive robustness value. The higher the minimum Euclidean distance between the~\ac{MRAV}'s trajectory and the region's boundaries, the greater the robustness. However, in~\eqref{subeq:avoidObostacles} and~\eqref{subeq:behind}, this relationship is inverted, with the~\ac{MRAV}'s presence in an obstacle region or a space behind the human operator resulting in a negative robustness value. To provide a specific example, let us consider the $\varphi_\mathrm{ws}$ predicate~\eqref{subeq:belongWorkspace}. This predicate can be mathematically expressed as follows:
\begin{equation}\label{eq:predicatesEqu}
    \resizebox{0.89\hsize}{!}{$%
    \rho_{\varphi_{\mathrm{ws}}} =  \min\limits_{k} \left (\min ( { \rho_{\bar{\varphi}^{(1)}} }, { \rho_{\underline{\varphi}^{(1)}} }, { \rho_{\bar{\varphi}^{(2)}} }, { \rho_{\underline{\varphi}^{(2)}} }, { \rho_{\bar{\varphi}^{(3)}} }, { \rho_{\underline{\varphi}^{(3)}} } ) \right ),
    $}%
\end{equation}
with
\begin{align*}
    { \rho_{\bar{\varphi}^{(j)}} } &= \bar{p}_\mathrm{ws}^{(j)} - p_k^{(j)},
    \quad
    { \rho_{\underline{\varphi}^{(j)}} } = p_k^{(j)} - \underline{p}_\mathrm{ws}^{(j)}. 
\end{align*}

Similarly, the robustness metric for the non-belonging predicate $\varphi_\mathrm{obs}$~\eqref{subeq:avoidObostacles} can be calculated by reversing the minimum distance values for each sample along the trajectory. Mathematically, this can be represented as:
\begin{equation}\label{eq:predicatesObsExplained}
    \resizebox{0.89\hsize}{!}{$%
    \rho_{\varphi_{\mathrm{obs}}} =  \min\limits_{k} \left (-\min ( { \rho_{\bar{\varphi}^{(1)}} }, { \rho_{\underline{\varphi}^{(1)}} }, { \rho_{\bar{\varphi}^{(2)}} }, { \rho_{\underline{\varphi}^{(2)}} }, { \rho_{\bar{\varphi}^{(3)}} }, { \rho_{\underline{\varphi}^{(3)}} } ) \right ),
    $}%
\end{equation}
where $\rho_{\bar{\varphi}^{(j)}}$ and $\rho_{\underline{\varphi}^{(j)}}$, with $j=\{1,2,3\}$, can be computed by replacing $\bar{\rho}^{(j)}_\mathrm{ws}$ and $\underline{\rho}^{(j)}_\mathrm{ws}$ in~\eqref{eq:predicatesEqu} with $\bar{\rho}^{(j)}_\mathrm{obs}$ and $\underline{\rho}^{(j)}_\mathrm{obs}$, respectively.

It is worth mentioning that the~\ac{NLP} problem is tackled by means of dynamic programming, which has a tendency to get trapped in local optima if the initial guess is not well-chosen~\cite{Bertsekas2012Book, CaiMacroeconomicDynamics2017}. Therefore, selecting an appropriate initial guess is of paramount importance. 



\subsection{Initial guess}
\label{sec:initialGuess}

An appropriate initial guess is essential to arrive at optimal solutions for the~\ac{STL} motion planner within a reasonable time frame, and also to prevent the solver from getting stuck during the search for a feasible solution. The strategy for obtaining this initial guess involves simplifying the original problem by abstracting it in order to yield an optimization problem with fewer constraints. Specifically, the initial guess takes into account the mission requirements that pertain to visiting all human operators, while noting the~\ac{MRAV} payload capacity and refilling operations ($\varphi_\mathrm{han}$, $\varphi_\mathrm{cap}$, and $\varphi_\mathrm{rs}$). However, it disregards the obstacle avoidance and ergonomy requirements ($\varphi_\mathrm{obs}$, $\varphi_\mathrm{ws}$, $\varphi_\mathrm{beh}$, and $\varphi_\mathrm{pr}$), as well as mission time intervals ($t_N$, $t_\mathrm{han}$ and $t_\mathrm{rs}$) as these constraints introduce the most complex nonlinearities and motion discontinuities in the problem. 

A graph-based representation, including connections between the human operators and the refilling stations, is employed to formulate an~\ac{ILP} problem that models a type of capacitated~\ac{VRP}~\cite{LaValleBook}. The resulting solution assigns the human operators (those they deliver a tool to) to the vehicle and provides a navigation sequence for the~\ac{MRAV}. 

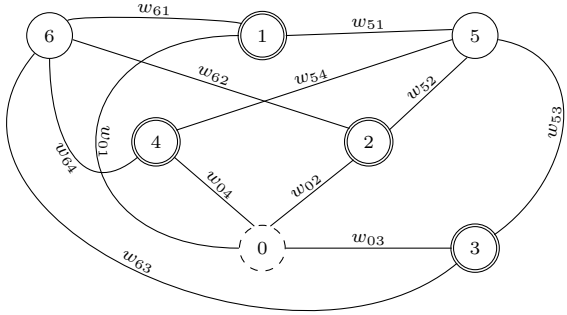
\begin{figure}[tb]
    \hspace{-4em}
    \centering
    \begin{tikzpicture}[node distance=20mm, on grid, auto, inner sep=4pt] 

    \node[draw, circle, double] (1) {\scriptsize{$4$}};  
    \node[draw, circle, double] (2) [above right of=1] {\scriptsize{$1$}}; 
    \node[draw, circle, dashed] (3) [below right of=1] {\scriptsize{$0$}}; 
    \node[draw, circle, double] (4) [above right of=3] {\scriptsize{$2$}}; 
    \node[draw, circle, double] (5) [below right of=4] {\scriptsize{$3$}}; 
    \node[draw, circle, solid] (6) [above right of=4] {\scriptsize{$5$}}; 
    \node[draw, circle, solid] (7) [above left of=1] {\scriptsize{$6$}}; 
    
    \draw[-] (2) to [out=180, in=180, looseness=2.3] node[midway, above, sloped, pos=0.5, yshift=-3pt] {\scriptsize{$w_{01}$}} (3);
    \draw[-] (4) to [out=230, in=70, looseness=0] node[midway, above, sloped, pos=0.5, yshift=-3pt] {\scriptsize{$w_{02}$}} (3);
    \draw[-] (3) to [out=0, in=180, looseness=0] node[midway, above, sloped, pos=0.5, yshift=-3pt] {\scriptsize{$w_{03}$}} (5);
    \draw[-] (3) to [out=110, in=-40, looseness=0] node[midway, above, sloped, pos=0.5, yshift=-3pt] {\scriptsize{$w_{04}$}} (1);
    
    \draw[-] (1) to [out=30, in=190, looseness=0] node[midway, above, sloped, pos=0.5, yshift=-3pt] {\scriptsize{$w_{54}$}} (6);
    \draw[-] (2) to [out=0, in=165, looseness=0] node[midway, above, sloped, pos=0.5, yshift=-3pt] {\scriptsize{$w_{51}$}} (6);
    \draw[-] (4) to [out=30, in=-110, looseness=0] node[midway, above, sloped, pos=0.5, yshift=-3pt] {\scriptsize{$w_{52}$}} (6);
    \draw[-] (5) to [out=35, in=-10, looseness=1.3] node[midway, above, sloped, pos=0.5, yshift=-3pt] {\scriptsize{$w_{53}$}} (6);

    \draw[-] (1) to [out=220, in=-90, looseness=1.45] node[midway, above, sloped, pos=0.5, yshift=-9pt] {\scriptsize{$w_{64}$}} (7);
    \draw[-] (2) to [out=150, in=40, looseness=0.15] node[midway, above, sloped, pos=0.5, yshift=-3pt] {\scriptsize{$w_{61}$}} (7);
    \draw[-] (5) to [out=220, in=-130, looseness=1.15] node[midway, above, sloped, pos=0.5, yshift=-3pt] {\scriptsize{$w_{63}$}} (7);
    \draw[-] (4) to [out=145, in=-10, looseness=0] node[midway, above, sloped, pos=0.5, yshift=-3pt] {\scriptsize{$w_{62}$}} (7);
    
    \end{tikzpicture} 
    \vspace{-3.65em}
    \caption{Instance of the graph $G$ assuming four human operators (double round nodes), two refilling stations (round solid nodes), and one vehicle. Arcs and weights $w_{ij}$ denote the~\acp{MRAV}'s path. The depot is depicted by a round dahsed node.}
    \label{fig:undirectedMultigraph}
\end{figure}

The graph used to formulate the~\ac{ILP} problem is illustrated in Figure~\ref{fig:undirectedMultigraph} and is characterized by the tuple $G = (\mathcal{V}, \mathcal{E}, \mathcal{W}, \mathcal{C})$. The set of vertices denoted by $\mathcal{V}$ is comprised of the human operators ($\mathcal{T}$), refilling stations ($\mathcal{R}$), and the depot ($\mathcal{O}$) where the~\ac{MRAV} is located at the initial time $t_0$. The number of elements in $\mathcal{T}$, $\mathcal{R}$, and $\mathcal{O}$ are represented by $\tau$, $r$, and $\delta$, respectively. The set of edges and their associated weights are represented by $\mathcal{E}$ and $\mathcal{W}$, respectively. Furthermore, $\mathcal{C}$ represents the vehicle maximum payload capacity. In terms of connectivity, all vertices in $\mathcal{T}$ are fully connected to every vertex in $\mathcal{R} \cup \mathcal{O}$. Formally, let $e_{ij} \in \mathcal{E}$ represent the edge connecting the vertices $i$ and $j$, with $\{i,j\} \in \mathcal{V}, i \neq j$ and $w_{ij} \in \mathcal{W}$ the weight associated with $e_{ij}$. Given the dynamic constraints for the~\ac{MRAV} $\bar{v}^{(j)}$ and $\bar{a}^{(j)}$, we model the edge weights using Euclidean distances, implying that $w_{ij} = w_{ji}$. Therefore, minimizing the mission time can be equivalent to reducing the distance traveled by the vehicle. 


To represent the number of times an edge is selected in the~\ac{ILP} solution, an integer variable $z_{ij} \in \mathbb{Z}_{\geq 0}$ is defined for each edge $e_{ij} \in \mathcal{E}$, and holds that $z_{ij} = z_{ji}$. This variable is used to specify the number of times the corresponding edge is selected in the~\ac{ILP} solution. Thus, $z_{ij}$ is limited to the set $\{0,1\}$ if $\{i,j\} \in \{ \mathcal{T},\mathcal{O} \}$ and $z_{ij}$ is limited to the set $\{0,1,2\}$ if $i \in \mathcal{R}$ and $j \in \mathcal{T}$. This limitation ensures that an edge between two human operators is never traversed twice and that the depot is only used as a starting point. This makes the proposed solution further suitable for wider scenarios where the~\ac{MRAV} can carry more than one tool at a time. Additionally, this allows for round trips between refilling stations and human operators in case there are no other tools to be delivered, i.e., $z_{ij} = 2$. The variable $z_{ij} = 0$ corresponds to non-traversed edges. By utilizing all of these defined variables, the~\ac{ILP} problem can be formulated as follows:
\begin{subequations}\label{eq:ILP}
    \begin{align}
        &\minimize_{z_{ij}}
        { \sum\limits_{ \{i,j\} \in \mathcal{V}, \, i \neq j} \hspace{-1em} w_{ij} \, z_{ij} } \label{subeq:objectiveFunction} \\
        %
        &\quad \;\; \text{s.t.} \;\, \hspace{-0.15cm} \sum\limits_{i \in \mathcal{V}, \, i \neq j} z_{ij} = 2, \; \forall j \in \mathcal{T}, \label{subeq:visitedOnce} \\ 
        %
        %
        &\qquad \;\,\;\;\;\; \sum\limits_{ i \in \mathcal{T} } \quad \hspace{-0.285em} z_{0i} = 1, \label{subeq:depotVisitedOnce} \\ 
        &\qquad \;\; \sum\limits_{\substack{ i \in \mathcal{T}, \, j \not\in \mathcal{T}}} \quad \hspace{-1.15em} z_{ij} \geq 2 h\hspace{-0.2em}\left(\mathcal{T}\right), \label{subeq:capacityAndSubtours} 
     \end{align}
\end{subequations}
where~\eqref{subeq:objectiveFunction} represents the objective function that encompasses the overall distance traversed by the~\ac{MRAV}. The constraints~\eqref{subeq:visitedOnce} mandate that each human operator is visited only once. Equation~\eqref{subeq:depotVisitedOnce} ensures that each~\ac{MRAV} begins its mission at its depot and does not return.
Constraints~\eqref{subeq:capacityAndSubtours} prevent the formation of tours that exceed the payload capacity of the~\ac{MRAV} or those which are not connected to a refilling station. This is achieved by using the function $h\hspace{-0.2em}\left(\mathcal{T}\right)$~\cite{LaValleBook} and dynamically adding these constraints as they are violated.

Once the optimal assignment for the~\ac{ILP} problem is obtained, motion primitives for the vehicle~\cite[eq.(2)]{SilanoRAL2021} are used to obtain a dynamically feasible trajectory that correspond to its optimal assignment. The methodology to compute these trajectories is described in~\cite{SilanoRAL2021} and is not covered in this paper for conciseness. In summary, the motion primitives are calculated by fixing rest-to-rest motion between operators, resulting in zero velocity and acceleration in these places, and imposing the minimum feasible time that allows for the desired maximum values of velocity $\bar{v}^{(j)}$ and acceleration $\bar{a}^{(j)}$ along the~\ac{MRAV}'s motion. The time intervals for handover $t_\mathrm{han}$ and refilling $t_\mathrm{rs}$ with the~\ac{MRAV} stopped in the corresponding regions are also taken into account for the trajectory.



\section{Simulation Results}
\label{sec:simulationResults}

To validate the proposed planning approach, numerical simulations were carried out in MATLAB. At this stage, the vehicle dynamics and trajectory tracking controller were not included in the simulations. The trajectories were then verified for feasibility in realistic simulations performed in Gazebo while exploiting the benefits of software-in-the-loop simulations~\cite{SilanoICUAS2021}. 

\begin{figure}[tb]
    \centering
    \adjincludegraphics[trim={{.1\width} {.03\height} {.1\width} {.33\height}}, clip, width=0.8\columnwidth]{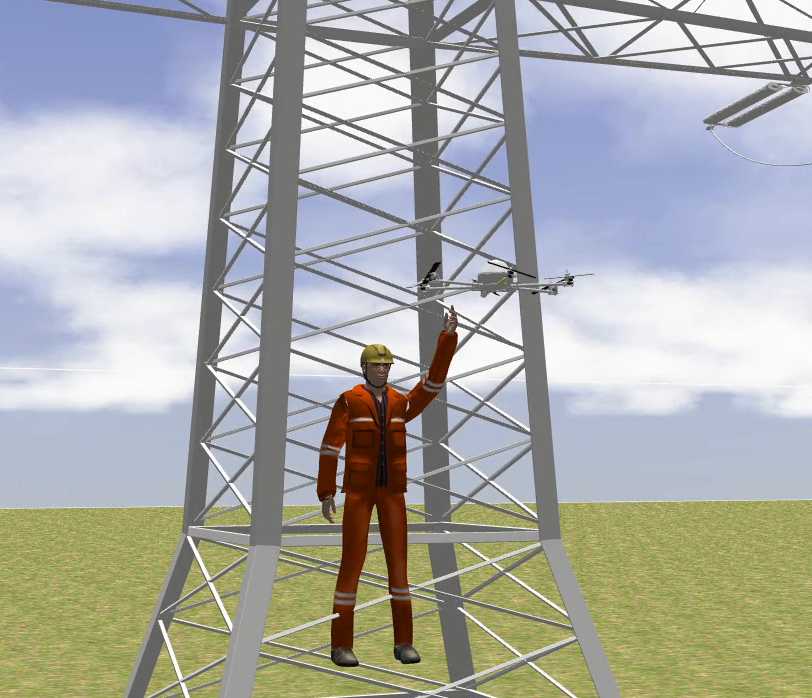}
    \vspace{-0.4em}
    \caption{A snapshot of the handover task in the Gazebo simulator featuring an~\ac{MRAV} and a human operator.}
    \label{fig:snapshotGazebo}
\end{figure}

The optimization method was implemented in MATLAB R2019b, with the~\ac{ILP} problem formulated using the CVX framework and the~\ac{STL} motion planner using the CasADi library and IPOPT as the solver. All simulations were run on a computer with an i7-8565U processor (\SI{1.80}{\giga\hertz}) and $32$GB of RAM running on Ubuntu 20.04. Figure~\ref{fig:snapshotGazebo} depicts a snapshot of the object handover task, while illustrative videos with the simulations are available at~\url{http://mrs.felk.cvut.cz/stl-ergonomy}.

The proposed planning strategy was evaluated using the object handover scenario outlined in Section~\ref{sec:problemDescription}. In particular, as shown in Figure~\ref{fig:scenarioSTL}, the simulation scenario consisted of a mock-up environment ($\SI{50}{\meter} \times \SI{20}{\meter} \times \SI{15}{\meter}$) with two human operators, one refilling station, and a single~\ac{MRAV}. The parameters and their corresponding values used to run the optimization problem are listed in Table~\ref{tab:tableParamters}. It is worth noting that the handover time $t_\mathrm{han}$ and the refilling time $t_\mathrm{rs}$ were set to short symbolic times to avoid analyzing trajectories with excessive waiting times. The heading angle of the~\ac{MRAV} was adjusted by aligning the vehicle with the direction of movement when moving towards the human operator. Once the~\ac{MRAV} reaches the operator, it is assumed that an onboard low-level controller, e.g.~\cite{Afifi2022ICRA, Corsini2022IROS}, handles the handover operation, thus adjusting the heading angle accordingly. 
The rectangular regions in which the drone was allowed to approach the operator were established taking into consideration the operators' heading, $\psi_\mathrm{ho1}$ and $\psi_\mathrm{ho2}$, as well as their preferred direction of approach ($\varphi_\mathrm{pr}$).

\begin{table}[tb]
    \centering
	\begin{adjustbox}{max width=1\columnwidth}
	\begin{tabular}{|c|c|c|}
        \hline
        \textbf{Parameter} & \textbf{Symbol} & \textbf{Value}
        \\
        \hline
        Payload capacity & $c$  & $\SI{1}{[-]}$ \\
        Max. velocity & $\bar{v}^{(j)}$ & $\SI{1.1}{[\meter\per\second]}$ \\
	Max. acceleration & $\bar{a}^{(j)}$ & $\SI{1.1} {[\meter\per\square\second]}$ \\
        Mission time & $t_N$ & $\SI{23}{[\second]}$ \\
        Handover time & $t_\mathrm{han}$ & $\SI{3}{[\second]}$ \\
        Refilling time & $t_\mathrm{rs}$ & $\SI{3}{[\second]}$ \\
        Sampling period & $T_s$ & $\SI{0.05}{[\second]}$ \\
        Number of samples & $N$ & $\SI{460}{[-]}$ \\
        Heading operator HO1 & $\psi_{\mathrm{ho1}}$ & $\si{\pi} \si{[\radian]}$ \\
        Heading operator HO2 & $\psi_{\mathrm{ho2}}$ & $\si{0} \si{[\radian]}$ \\
	\hline
	\end{tabular}
	\end{adjustbox}
        \vspace{-0.4em}
	\caption{Parameter values for the optimization problem.}
	\label{tab:tableParamters}
\end{table}

\begin{figure}[tb]
    \centering
    \adjincludegraphics[trim={{.075\width} {.05\height} {.08\width} {.08\height}}, clip, width=\columnwidth]{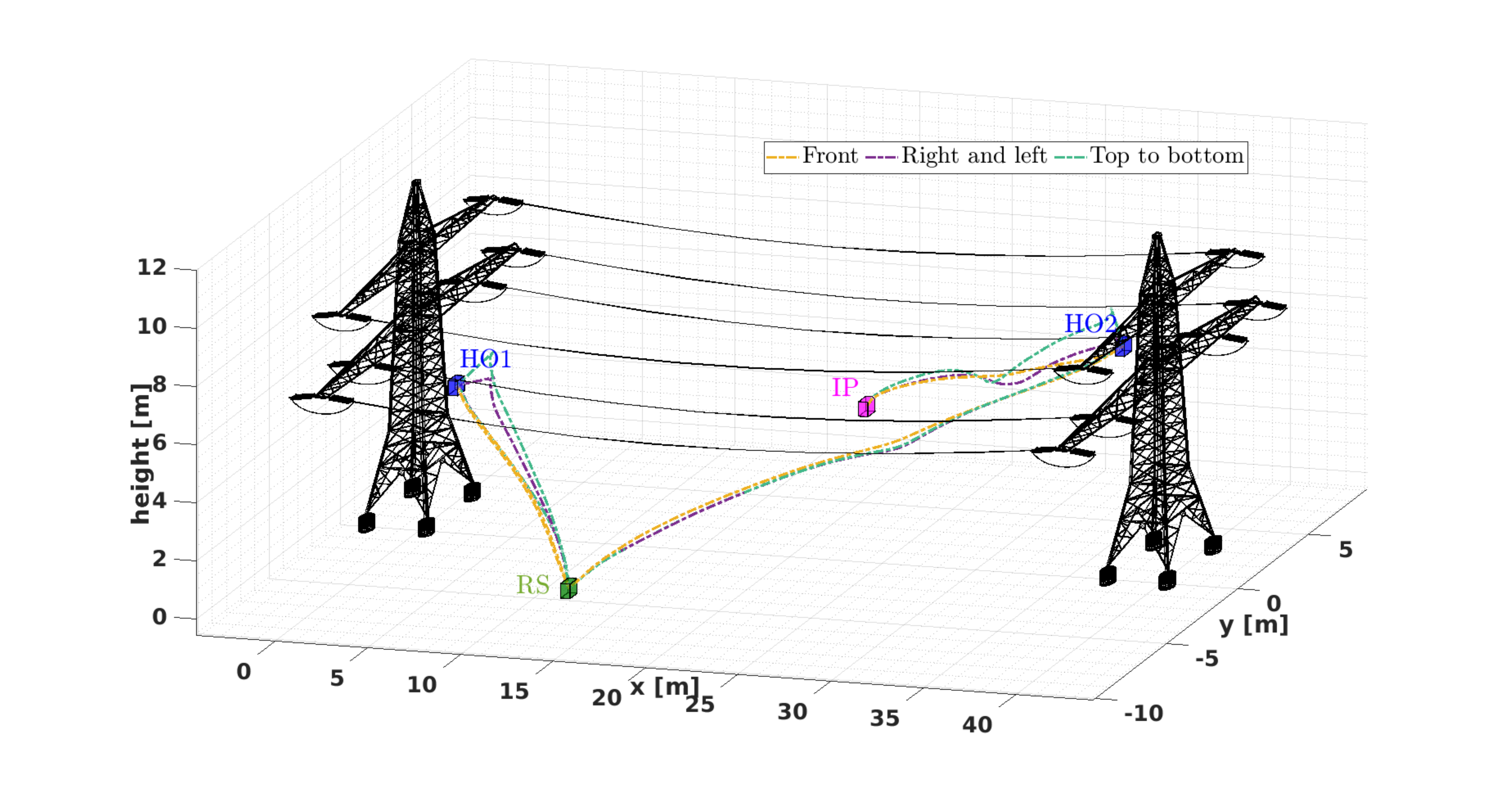}
    \vspace{-1.75em}
    \caption{Power line scenario. The trajectories considering different operators' preferred approach directions, such as front, left and right, and top to bottom, and are depicted in yellow, purple, and green, respectively.}
    \label{fig:planner_base}
\end{figure}

Figure~\ref{fig:planner_base} illustrates the planned trajectories along with the power towers and cables, human operators, and refilling station. The figure compares the trajectories computed by taking into account the preferred approach directions of the operators, including front, right and left, and top to bottom. The towers are $\SI{15}{\meter}$ tall and are positioned $\SI{40}{\meter}$ apart. The optimization problem was solved in $\SI{130}{\second}$ and it took $\SI{1}{\second}$ to find an initial guess solution. Note that, even though the initial guess solution retrieved by solving the~\ac{ILP} problem is the same, the~\ac{STL} optimizer rearranges the trajectory to maximize the robustness score. This basic scenario illustrates the decoupling of the final~\ac{STL} optimization from the~\ac{ILP} initial solution, which provides added versatility to the proposed human-robot collaboration planner. Additionally, it highlights the ease with which high-level ergonomic requirements can be integrated into the problem formulation. However, as the complexity increases, beginning with an initial solution far from the overall problem solution can lead to the~\ac{STL} optimizer becoming trapped in a local optimum.

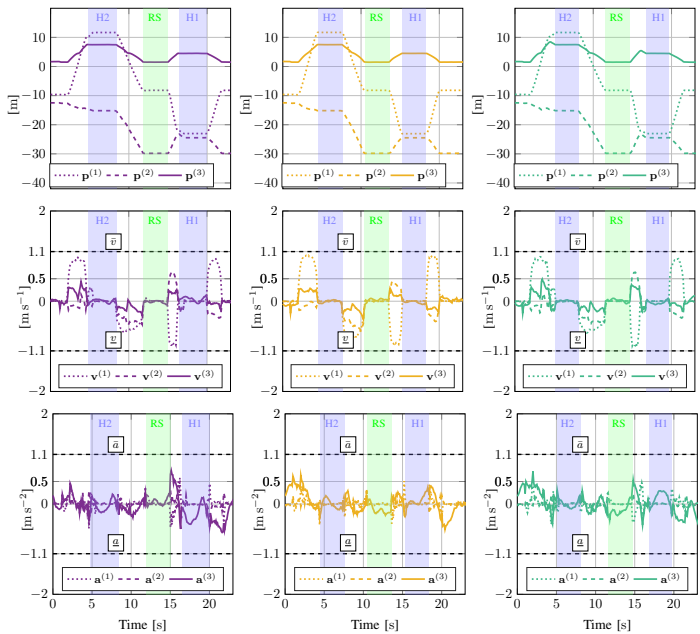
\begin{figure}[tb]
    \centering	
    \input{matlabPlots/experiments_plot.tex}
    \vspace{-1.65mm}
    \caption{Position, linear velocity, and acceleration. The trajectories are shown for the different operators' preferred approach directions, including left and right (left), front (middle), and top to bottom (right). The time frames for handover and refilling are indicated using blue and green colors, respectively.}
    \label{fig:scenario_9RealWorldExp}
\end{figure}

Figure~\ref{fig:scenario_9RealWorldExp} demonstrates that the planned trajectories comply with the mission requirements. As can be observed from the graph, the vehicle velocity and acceleration stay within the permissible bounds ($[\underline{v}^{(j)}, \bar{v}^{(j)}]$ and $[\underline{a}^{(j)}, \bar{a}^{(j)}]$). For simplicity, the velocity and acceleration bounds are considered to be symmetric, i.e., $\lvert \underline{v}^{(j)} \rvert = \lvert \bar{v}^{(j)} \rvert$ and $\lvert \underline{a}^{(j)} \rvert = \lvert \bar{a}^{(j)} \rvert$. In addition, the time frames for both the handover and refilling operations are presented, highlighting how the user's preferences directly reflect on the~\ac{MRAV} motion.

\section{Conclusions}
\label{sec:conclusions}

In this paper, a motion planning framework was introduced to enhance ergonomic human-robot collaboration for an MRAV with payload capacity limitations and dynamic constraints. The proposed method employs~\ac{STL} specifications to generate trajectories that comply with mission requirements, including safety, ergonomics, and mission time. This work extends a previous motion planner~\cite{SilanoRAL2021, CaballeroAR2022} by introducing a method for tackling the nonlinear non-convex nature of the optimization problem through the use of an~\ac{ILP} approach. This approach utilizes a simplified version of mission specifications to provide a feasible initial solution for the~\ac{STL} framework to facilitate convergence of the algorithm. Numerical analyses in MATLAB and realistic simulations in Gazebo demonstrated the effectiveness of the proposed approach. Future work includes incorporating human operator fatigue into the problem formulation by using weights associated with Boolean and temporal operators to modulate the robustness. Additionally, research will be conducted on the use of conflicting temporal logic specifications and other types of temporal logic languages to adapt the framework for dynamic environments.



\bibliographystyle{IEEEtran}
\bibliography{bib_short}

\end{document}

%% file: matlabPlots/experiments_plot.tex
\vspace{0.50em}
\hspace{-0.625cm}
\begin{subfigure}{0.30\columnwidth}
    \centering
    \scalebox{0.52}{
    \begin{tikzpicture}
    \begin{axis}[%
    width=1.8119in,%
    height=1.8183in,%
    at={(0.758in,0.481in)},%
    scale only axis,%
    xmin=0,%
    xmax=23,%
    ymax=20,%
    ymin=-42,%
    xmajorgrids,%
    ymajorgrids,%
    ylabel style={yshift=-0.285cm, xshift=-0.15cm}, 
    extra x ticks={33.50,38.50,50.50,55.50,67.00,79.00,115.00,120.00,131.00,143.00},%
    extra x tick labels={ , ,  , },%
    ylabel={[\si{\meter}]},%
    ytick={10,0,...,-40},%
    xticklabels={ , , },%
    axis background/.style={fill=white},%
    legend style={at={(0.45,0.14)},anchor=north,legend cell align=left,draw=none,legend 
    columns=-1,align=left,draw=white!15!black}
    ]
    \addplot [color=myPurple, dotted, line width=1.25pt] 
        file{matlabPlots/experiments/ergonomy_leftRight/position_UAV1_x.txt};%
    \addplot [color=myPurple, dashed, line width=1.25pt] 
        file{matlabPlots/experiments/ergonomy_leftRight/position_UAV1_y.txt};%
    \addplot [color=myPurple, solid, line width=1.25pt] 
        file{matlabPlots/experiments/ergonomy_leftRight/position_UAV1_z.txt};%
    \fill[blue!50,nearly transparent] (axis cs:{4.85,-42}) -- (axis cs:{4.85,20}) -- (axis 
    cs:{8.45,20}) -- (axis cs:{8.45,-42}) -- cycle;
    \fill[green!50,nearly transparent] (axis cs:{11.90,-42}) -- (axis cs:{11.90,20}) -- (axis 
    cs:{15.00,20}) -- (axis cs:{15.00,-42}) -- cycle;
    \fill[blue!50,nearly transparent] (axis cs:{16.50,-42}) -- (axis cs:{16.50,20}) -- (axis 
    cs:{20.00,20}) -- (axis cs:{20.00,-42}) -- cycle;
    \legend{\footnotesize{$\mathbf{p}^{(1)}$}, \footnotesize{$\mathbf{p}^{(2)}$}, 
    \footnotesize{$\mathbf{p}^{(3)}$}};%
    \node at (rel axis cs: 0.29,0.95) 
    {\shortstack[l]{\footnotesize{\color{blue!50}{H2}}}};
    \node at (rel axis cs: 0.58,0.95) 
    {\shortstack[l]{\footnotesize{\color{green!100!black!100}{RS}}}};
    \node at (rel axis cs: 0.79,0.95) 
    {\shortstack[l]{\footnotesize{\color{blue!50}{H1}}}};
    \end{axis}
    \end{tikzpicture}
    }
\end{subfigure}
\hspace{0.25cm}
\begin{subfigure}{0.30\columnwidth}
    \centering
    \scalebox{0.52}{
    \begin{tikzpicture}
    \begin{axis}[%
    width=1.8119in,%
    height=1.8183in,%
    at={(0.758in,0.481in)},%
    scale only axis,%
    xmin=0,%
    xmax=23,%
    ymax=20,%
    ymin=-42,%
    xmajorgrids,%
    ymajorgrids,%
    ylabel style={yshift=-0.285cm, xshift=-0.15cm}, 
    extra x ticks={33.50,38.50,50.50,55.50,67.00,79.00,115.00,120.00,131.00,143.00},%
    extra x tick labels={ , ,  , },%
    ylabel={[\si{\meter}]},%
    ytick={10,0,...,-40},%
    xticklabels={ , , },%
    axis background/.style={fill=white},%
    legend style={at={(0.45,0.14)},anchor=north,legend cell align=left,draw=none,legend 
    columns=-1,align=left,draw=white!15!black}
    ]
    \addplot [color=myYellow, dotted, line width=1.25pt] 
    file{matlabPlots/experiments/ergonomy_LP/position_UAV1_x.txt};%
    \addplot [color=myYellow, dashed, line width=1.25pt] 
    file{matlabPlots/experiments/ergonomy_LP/position_UAV1_y.txt};%
    \addplot [color=myYellow, solid, line width=1.25pt] 
    file{matlabPlots/experiments/ergonomy_LP/position_UAV1_z.txt};%
    \fill[blue!50,nearly transparent] (axis cs:{4.55,-42}) -- (axis cs:{4.55,20}) -- (axis 
    cs:{7.65,20}) -- (axis cs:{7.65,-42}) -- cycle;
    \fill[green!50,nearly transparent] (axis cs:{10.50,-42}) -- (axis cs:{10.50,20}) -- (axis 
    cs:{13.60,20}) -- (axis cs:{13.60,-42}) -- cycle;
    \fill[blue!50,nearly transparent] (axis cs:{15.35,-42}) -- (axis cs:{15.35,20}) -- (axis 
    cs:{18.30,20}) -- (axis cs:{18.30,-42}) -- cycle;
    \legend{\footnotesize{$\mathbf{p}^{(1)}$}, \footnotesize{$\mathbf{p}^{(2)}$}, 
    \footnotesize{$\mathbf{p}^{(3)}$}};%
    \node at (rel axis cs: 0.26,0.95) 
    {\shortstack[l]{\footnotesize{\color{blue!50}{H2}}}};
    \node at (rel axis cs: 0.52,0.95) 
    {\shortstack[l]{\footnotesize{\color{green!100!black!100}{RS}}}};
    \node at (rel axis cs: 0.74,0.95) 
    {\shortstack[l]{\footnotesize{\color{blue!50}{H1}}}};
    \end{axis}
    \end{tikzpicture}
     }
\end{subfigure}
\hspace{0.25cm}
\begin{subfigure}{0.30\columnwidth}
    \centering
    \scalebox{0.52}{
    \begin{tikzpicture}
    \begin{axis}[%
    width=1.8119in,%
    height=1.8183in,%
    at={(0.758in,0.481in)},%
    scale only axis,%
    xmin=0,%
    xmax=23,%
    ymax=20,%
    ymin=-42,%
    xmajorgrids,%
    ymajorgrids,%
    ylabel style={yshift=-0.285cm, xshift=-0.15cm}, 
    extra x ticks={33.50,38.50,50.50,55.50,67.00,79.00,115.00,120.00,131.00,143.00},%
    extra x tick labels={ , ,  , },%
    ylabel={[\si{\meter}]},%
    ytick={10,0,...,-40},%
    xticklabels={ , , },%
    axis background/.style={fill=white},%
    legend style={at={(0.45,0.14)},anchor=north,legend cell align=left,draw=none,legend 
    columns=-1,align=left,draw=white!15!black}
    ]
    \addplot [color=myGreen, dotted, line width=1.25pt] 
    file{matlabPlots/experiments/ergonomy_topDown/position_UAV1_x.txt};%
    \addplot [color=myGreen, dashed, line width=1.25pt] 
    file{matlabPlots/experiments/ergonomy_topDown/position_UAV1_y.txt};%
    \addplot [color=myGreen, solid, line width=1.25pt] 
    file{matlabPlots/experiments/ergonomy_topDown/position_UAV1_z.txt};%
   \fill[blue!50,nearly transparent] (axis cs:{5.10,-42}) -- (axis cs:{5.10,20}) -- (axis 
    cs:{8.15,20}) -- (axis cs:{8.15,-42}) -- cycle;
    \fill[green!50,nearly transparent] (axis cs:{11.60,-42}) -- (axis cs:{11.60,20}) -- (axis 
    cs:{14.70,20}) -- (axis cs:{14.70,-42}) -- cycle;
    \fill[blue!50,nearly transparent] (axis cs:{16.80,-42}) -- (axis cs:{16.80,20}) -- (axis 
    cs:{19.70,20}) -- (axis cs:{19.70,-42}) -- cycle;
    \legend{\footnotesize{$\mathbf{p}^{(1)}$}, \footnotesize{$\mathbf{p}^{(2)}$}, 
    \footnotesize{$\mathbf{p}^{(3)}$}};%
    \node at (rel axis cs: 0.28,0.95) 
    {\shortstack[l]{\footnotesize{\color{blue!50}{H2}}}};
    \node at (rel axis cs: 0.565,0.95) 
    {\shortstack[l]{\footnotesize{\color{green!100!black!100}{RS}}}};
    \node at (rel axis cs: 0.80,0.95) 
    {\shortstack[l]{\footnotesize{\color{blue!50}{H1}}}};
    \end{axis}
    \end{tikzpicture}
    }
\end{subfigure}
\\
\vspace{0.05cm}
\hspace{-0.375cm}
\begin{subfigure}{0.30\columnwidth}
    \centering
    \scalebox{0.52}{
    \begin{tikzpicture} 
    \begin{axis}[%
    width=1.8119in,%
    height=1.8183in,%
    at={(0.758in,0.481in)},%
    scale only axis,%
    xmin=0,%
    xmax=23,%
    ymax=2,%
    ymin=-2,%
    xmajorgrids,%
    ymajorgrids,%
    ylabel style={yshift=-0.555cm, xshift=-0.15cm}, 
    extra x ticks={33.50,38.50,50.50,55.50,67.00,79.00,115.00,120.00,131.00,143.00},%
    extra x tick labels={ , ,  , },%
    ylabel={[\si{\meter\per\second}]},%
    ytick={-2,-1.1,0.5,0,0.5,1.1,2},%
    xticklabels={ , , },%
    axis background/.style={fill=white},%
    legend style={at={(0.5,0.15)},anchor=north,legend cell align=left,draw=none,legend 
        columns=-1,align=left,draw=white!15!black}
    ]
    \addplot [color=myPurple, dotted, line width=1.25pt] 
    file{matlabPlots/experiments/ergonomy_leftRight/velocity_UAV1_x.txt};%
    \addplot [color=myPurple, dashed, line width=1.25pt] 
    file{matlabPlots/experiments/ergonomy_leftRight/velocity_UAV1_y.txt};%
    \addplot [color=myPurple, solid, line width=1.25pt] 
    file{matlabPlots/experiments/ergonomy_leftRight/velocity_UAV1_z.txt};%
    \draw [line width=1.00pt, dashed] (0, 90) -- (280, 90); 
    \draw [line width=1.00pt, dashed] (0, 310) -- (280, 310); 
    \fill[blue!50,nearly transparent] (axis cs:{4.85,-42}) -- (axis cs:{4.85,20}) -- (axis 
    cs:{8.45,20}) -- (axis cs:{8.45,-42}) -- cycle;
    \fill[green!50,nearly transparent] (axis cs:{11.90,-42}) -- (axis cs:{11.90,20}) -- (axis 
    cs:{15.00,20}) -- (axis cs:{15.00,-42}) -- cycle;
    \fill[blue!50,nearly transparent] (axis cs:{16.50,-42}) -- (axis cs:{16.50,20}) -- (axis 
    cs:{20.00,20}) -- (axis cs:{20.00,-42}) -- cycle;
    \legend{\footnotesize{$\mathbf{v}^{(1)}$}, \footnotesize{$\mathbf{v}^{(2)}$}, 
    \footnotesize{$\mathbf{v}^{(3)}$}};%
    \node [draw, fill=white] at (rel axis cs: 0.35,0.83) 
    {\shortstack[l]{\footnotesize{$\bar{v}$}}};
    \node [draw, fill=white] at (rel axis cs: 0.35,0.285) 
    {\shortstack[l]{\footnotesize{$\underline{v}$}}};
    \node at (rel axis cs: 0.29,0.95) 
    {\shortstack[l]{\footnotesize{\color{blue!50}{H2}}}};
    \node at (rel axis cs: 0.58,0.95) 
    {\shortstack[l]{\footnotesize{\color{green!100!black!100}{RS}}}};
    \node at (rel axis cs: 0.79,0.95) 
    {\shortstack[l]{\footnotesize{\color{blue!50}{H1}}}};
    \end{axis}
    \end{tikzpicture}
    }
\end{subfigure}
\hspace{0.25cm}
\begin{subfigure}{0.30\columnwidth}
    \centering
    \scalebox{0.52}{
    \begin{tikzpicture}
    \begin{axis}[%
    width=1.8119in,%
    height=1.8183in,%
    at={(0.758in,0.481in)},%
    scale only axis,%
    xmin=0,%
    xmax=23,%
    ymax=2.0,%
    ymin=-2.0,%
    xmajorgrids,%
    ymajorgrids,%
    ylabel style={yshift=-0.555cm, xshift=-0.15cm}, 
    extra x ticks={33.50,38.50,50.50,55.50,67.00,79.00,115.00,120.00,131.00,143.00},%
    extra x tick labels={ , ,  , },%
    ylabel={[\si{\meter\per\second}]},%
    ytick={-2,-1.1,0.5,0,0.5,1.1,2},%
    xticklabels={ , , },%
    axis background/.style={fill=white},%
    legend style={at={(0.5,0.15)},anchor=north,legend cell align=left,draw=none,legend 
        columns=-1,align=left,draw=white!15!black}
    ]
    \addplot [color=myYellow, dotted, line width=1.25pt] 
    file{matlabPlots/experiments/ergonomy_LP/velocity_UAV1_x.txt};%
    \addplot [color=myYellow, dashed, line width=1.25pt] 
    file{matlabPlots/experiments/ergonomy_LP/velocity_UAV1_y.txt};%
    \addplot [color=myYellow, solid, line width=1.25pt] 
    file{matlabPlots/experiments/ergonomy_LP/velocity_UAV1_z.txt};%
    \draw [line width=1.00pt, dashed] (0, 90) -- (280, 90); 
    \draw [line width=1.00pt, dashed] (0, 310) -- (280, 310); 
    \fill[blue!50,nearly transparent] (axis cs:{4.55,-42}) -- (axis cs:{4.55,20}) -- (axis 
    cs:{7.65,20}) -- (axis cs:{7.65,-42}) -- cycle;
    \fill[green!50,nearly transparent] (axis cs:{10.50,-42}) -- (axis cs:{10.50,20}) -- (axis 
    cs:{13.60,20}) -- (axis cs:{13.60,-42}) -- cycle;
    \fill[blue!50,nearly transparent] (axis cs:{15.35,-42}) -- (axis cs:{15.35,20}) -- (axis 
    cs:{18.30,20}) -- (axis cs:{18.30,-42}) -- cycle;
    \legend{\footnotesize{$\mathbf{v}^{(1)}$}, \footnotesize{$\mathbf{v}^{(2)}$}, 
        \footnotesize{$\mathbf{v}^{(3)}$}};%
    \node [draw, fill=white] at (rel axis cs: 0.35,0.83) 
    {\shortstack[l]{\footnotesize{$\bar{v}$}}};
    \node [draw, fill=white] at (rel axis cs: 0.35,0.285) 
    {\shortstack[l]{\footnotesize{$\underline{v}$}}};
    \node at (rel axis cs: 0.26,0.95) 
    {\shortstack[l]{\footnotesize{\color{blue!50}{H2}}}};
    \node at (rel axis cs: 0.52,0.95) 
    {\shortstack[l]{\footnotesize{\color{green!100!black!100}{RS}}}};
    \node at (rel axis cs: 0.74,0.95) 
    {\shortstack[l]{\footnotesize{\color{blue!50}{H1}}}};
    \end{axis}
    \end{tikzpicture}
}
\end{subfigure}
\hspace{0.25cm}
\begin{subfigure}{0.30\columnwidth}
    \centering
    \scalebox{0.52}{
    \begin{tikzpicture}
    \begin{axis}[%
    width=1.8119in,%
    height=1.8183in,%
    at={(0.758in,0.481in)},%
    scale only axis,%
    xmin=0,%
    xmax=23,%
    ymax=2.0,%
    ymin=-2.0,%
    xmajorgrids,%
    ymajorgrids,%
    ylabel style={yshift=-0.555cm, xshift=-0.15cm}, 
    extra x ticks={33.50,38.50,50.50,55.50,67.00,79.00,115.00,120.00,131.00,143.00},%
    extra x tick labels={ , ,  , },%
    ylabel={[\si{\meter\per\second}]},%
    ytick={-2,-1.1,0.5,0,0.5,1.1,2},%
    xticklabels={ , , },%
    axis background/.style={fill=white},%
    legend style={at={(0.5,0.15)},anchor=north,legend cell align=left,draw=none,legend 
        columns=-1,align=left,draw=white!15!black}
    ]
    \addplot [color=myGreen, dotted, line width=1.25pt] 
    file{matlabPlots/experiments/ergonomy_topDown/velocity_UAV1_x.txt};%
    \addplot [color=myGreen, dashed, line width=1.25pt] 
    file{matlabPlots/experiments/ergonomy_topDown/velocity_UAV1_y.txt};%
    \addplot [color=myGreen, solid, line width=1.25pt] 
    file{matlabPlots/experiments/ergonomy_topDown/velocity_UAV1_z.txt};%
    \draw [line width=1.00pt, dashed] (0, 90) -- (280, 90); 
    \draw [line width=1.00pt, dashed] (0, 310) -- (280, 310); 
    \fill[blue!50,nearly transparent] (axis cs:{5.10,-42}) -- (axis cs:{5.10,20}) -- (axis 
    cs:{8.15,20}) -- (axis cs:{8.15,-42}) -- cycle;
    \fill[green!50,nearly transparent] (axis cs:{11.60,-42}) -- (axis cs:{11.60,20}) -- (axis 
    cs:{14.70,20}) -- (axis cs:{14.70,-42}) -- cycle;
    \fill[blue!50,nearly transparent] (axis cs:{16.80,-42}) -- (axis cs:{16.80,20}) -- (axis 
    cs:{19.70,20}) -- (axis cs:{19.70,-42}) -- cycle;
    \legend{\footnotesize{$\mathbf{v}^{(1)}$}, \footnotesize{$\mathbf{v}^{(2)}$}, 
        \footnotesize{$\mathbf{v}^{(3)}$}};%
    \node [draw, fill=white] at (rel axis cs: 0.35,0.83) 
    {\shortstack[l]{\footnotesize{$\bar{v}$}}};
    \node [draw, fill=white] at (rel axis cs: 0.35,0.285) 
    {\shortstack[l]{\footnotesize{$\underline{v}$}}};
    \node at (rel axis cs: 0.28,0.95) 
    {\shortstack[l]{\footnotesize{\color{blue!50}{H2}}}};
    \node at (rel axis cs: 0.565,0.95) 
    {\shortstack[l]{\footnotesize{\color{green!100!black!100}{RS}}}};
    \node at (rel axis cs: 0.80,0.95) 
    {\shortstack[l]{\footnotesize{\color{blue!50}{H1}}}};
    \end{axis}
    \end{tikzpicture}
    }
\end{subfigure}
\\
\vspace{0.05cm}
\hspace{-0.31cm}
\begin{subfigure}{0.30\columnwidth}
    \centering
    \scalebox{0.52}{
    \begin{tikzpicture} 
    \begin{axis}[%
    width=1.8119in,%
    height=1.8183in,%
    at={(0.758in,0.481in)},%
    scale only axis,%
    xmin=0,%
    xmax=23,%
    ymax=2.0,%
    ymin=-2.0,%
    xmajorgrids,%
    ymajorgrids,%
    ylabel style={yshift=-0.555cm, xshift=-0.15cm}, 
    xlabel={Time [\si{\second}]},%
    extra x ticks={33.50,38.50,50.50,55.50,67.00,79.00,115.00,120.00,131.00,143.00},%
    extra x tick labels={ , ,  , },%
    xtick={0,5,10,15,20},%
    ylabel={[\si{\meter\per\second\squared}]},%
    ytick={-2,-1.1,0.5,0,0.5,1.1,2},%
    axis background/.style={fill=white},%
    legend style={at={(0.5,0.15)},anchor=north,legend cell align=left,draw=none,legend 
        columns=-1,align=left,draw=white!15!black}
    ]
    \addplot [color=myPurple, dotted, line width=1.25pt] 
    file{matlabPlots/experiments/ergonomy_leftRight/acceleration_UAV1_x.txt};%
    \addplot [color=myPurple, dashed, line width=1.25pt] 
    file{matlabPlots/experiments/ergonomy_leftRight/acceleration_UAV1_y.txt};%
    \addplot [color=myPurple, solid, line width=1.25pt] 
    file{matlabPlots/experiments/ergonomy_leftRight/acceleration_UAV1_z.txt};%
    \draw [line width=1.00pt, dashed] (0, 90) -- (280, 90); 
    \draw [line width=1.00pt, dashed] (0, 310) -- (280, 310); 
    \fill[blue!50,nearly transparent] (axis cs:{4.85,-42}) -- (axis cs:{4.85,20}) -- (axis 
    cs:{8.45,20}) -- (axis cs:{8.45,-42}) -- cycle;
    \fill[green!50,nearly transparent] (axis cs:{11.90,-42}) -- (axis cs:{11.90,20}) -- (axis 
    cs:{15.00,20}) -- (axis cs:{15.00,-42}) -- cycle;
    \fill[blue!50,nearly transparent] (axis cs:{16.50,-42}) -- (axis cs:{16.50,20}) -- (axis 
    cs:{20.00,20}) -- (axis cs:{20.00,-42}) -- cycle;
    \legend{\footnotesize{$\mathbf{a}^{(1)}$}, \footnotesize{$\mathbf{a}^{(2)}$}, 
        \footnotesize{$\mathbf{a}^{(3)}$}};%
    \node [draw, fill=white] at (rel axis cs: 0.35,0.83) 
    {\shortstack[l]{\footnotesize{$\bar{a}$}}};
    \node [draw, fill=white] at (rel axis cs: 0.35,0.285) 
    {\shortstack[l]{\footnotesize{$\underline{a}$}}};
    \node at (rel axis cs: 0.29,0.95) 
    {\shortstack[l]{\footnotesize{\color{blue!50}{H2}}}};
    \node at (rel axis cs: 0.58,0.95) 
    {\shortstack[l]{\footnotesize{\color{green!100!black!100}{RS}}}};
    \node at (rel axis cs: 0.79,0.95) 
    {\shortstack[l]{\footnotesize{\color{blue!50}{H1}}}};
    \end{axis}
    \end{tikzpicture}
    }
\end{subfigure}
\hspace{0.25cm}
\begin{subfigure}{0.30\columnwidth}
    \centering
    \scalebox{0.52}{
    \begin{tikzpicture}
    \begin{axis}[%
    width=1.8119in,%
    height=1.8183in,%
    at={(0.758in,0.481in)},%
    scale only axis,%
    xmin=0,%
    xmax=23,%
    ymax=2.0,%
    ymin=-2.0,%
    xmajorgrids,%
    ymajorgrids,%
    ylabel style={yshift=-0.555cm, xshift=-0.15cm}, 
    xlabel={Time [\si{\second}]},%
    extra x ticks={33.50,38.50,50.50,55.50,67.00,79.00,115.00,120.00,131.00,143.00},%
    extra x tick labels={ , ,  , },%
    xtick={0,5,10,15,20},%
    ylabel={[\si{\meter\per\second\squared}]},%
    ytick={-2,-1.1,0.5,0,0.5,1.1,2},%
    axis background/.style={fill=white},%
    legend style={at={(0.5,0.15)},anchor=north,legend cell align=left,draw=none,legend 
        columns=-1,align=left,draw=white!15!black}
    ]
    \addplot [color=myYellow, dotted, line width=1.25pt] 
    file{matlabPlots/experiments/ergonomy_LP/acceleration_UAV1_x.txt};%
    \addplot [color=myYellow, dashed, line width=1.25pt] 
    file{matlabPlots/experiments/ergonomy_LP/acceleration_UAV1_y.txt};%
    \addplot [color=myYellow, solid, line width=1.25pt] 
    file{matlabPlots/experiments/ergonomy_LP/acceleration_UAV1_z.txt};%
    \draw [line width=1.00pt, dashed] (0, 90) -- (280, 90); 
    \draw [line width=1.00pt, dashed] (0, 310) -- (280, 310); 
    \fill[blue!50,nearly transparent] (axis cs:{4.55,-42}) -- (axis cs:{4.55,20}) -- (axis 
    cs:{7.65,20}) -- (axis cs:{7.65,-42}) -- cycle;
    \fill[green!50,nearly transparent] (axis cs:{10.50,-42}) -- (axis cs:{10.50,20}) -- (axis 
    cs:{13.60,20}) -- (axis cs:{13.60,-42}) -- cycle;
    \fill[blue!50,nearly transparent] (axis cs:{15.35,-42}) -- (axis cs:{15.35,20}) -- (axis 
    cs:{18.30,20}) -- (axis cs:{18.30,-42}) -- cycle;
    \legend{\footnotesize{$\mathbf{a}^{(1)}$}, \footnotesize{$\mathbf{a}^{(2)}$}, 
        \footnotesize{$\mathbf{a}^{(3)}$}};%
    \node [draw, fill=white] at (rel axis cs: 0.35,0.83) 
    {\shortstack[l]{\footnotesize{$\bar{a}$}}};
    \node [draw, fill=white] at (rel axis cs: 0.35,0.285) 
    {\shortstack[l]{\footnotesize{$\underline{a}$}}};
    \node at (rel axis cs: 0.26,0.95) 
    {\shortstack[l]{\footnotesize{\color{blue!50}{H2}}}};
    \node at (rel axis cs: 0.52,0.95) 
    {\shortstack[l]{\footnotesize{\color{green!100!black!100}{RS}}}};
    \node at (rel axis cs: 0.74,0.95) 
    {\shortstack[l]{\footnotesize{\color{blue!50}{H1}}}};
    \end{axis}
    \end{tikzpicture}
    }
\end{subfigure}
\hspace{0.25cm}
\begin{subfigure}{0.30\columnwidth}
    \centering
    \scalebox{0.52}{
    \begin{tikzpicture}
    \begin{axis}[%
    width=1.8119in,%
    height=1.8183in,%
    at={(0.758in,0.481in)},%
    scale only axis,%
    xmin=0,%
    xmax=23,%
    ymax=2.0,%
    ymin=-2.0,%
    xmajorgrids,%
    ymajorgrids,%
    ylabel style={yshift=-0.555cm, xshift=-0.15cm}, 
    xlabel={Time [\si{\second}]},%
    extra x ticks={33.50,38.50,50.50,55.50,67.00,79.00,115.00,120.00,131.00,143.00},%
    xtick={0,5,10,15,20},%
    extra x tick labels={ , ,  , },%
    ylabel={[\si{\meter\per\second\squared}]},%
    ytick={-2,-1.1,0.5,0,0.5,1.1,2},%
    axis background/.style={fill=white},%
    legend style={at={(0.5,0.15)},anchor=north,legend cell align=left,draw=none,legend 
        columns=-1,align=left,draw=white!15!black}
    ]
    \addplot [color=myGreen, dotted, line width=1.25pt] 
    file{matlabPlots/experiments/ergonomy_topDown/acceleration_UAV1_x.txt};%
    \addplot [color=myGreen, dashed, line width=1.25pt] 
    file{matlabPlots/experiments/ergonomy_topDown/acceleration_UAV1_y.txt};%
    \addplot [color=myGreen, solid, line width=1.25pt] 
    file{matlabPlots/experiments/ergonomy_topDown/acceleration_UAV1_z.txt};%
    \draw [line width=1.00pt, dashed] (0, 90) -- (280, 90); 
    \draw [line width=1.00pt, dashed] (0, 310) -- (280, 310); 
    \fill[blue!50,nearly transparent] (axis cs:{5.10,-42}) -- (axis cs:{5.10,20}) -- (axis 
    cs:{8.15,20}) -- (axis cs:{8.15,-42}) -- cycle;
    \fill[green!50,nearly transparent] (axis cs:{11.60,-42}) -- (axis cs:{11.60,20}) -- (axis 
    cs:{14.70,20}) -- (axis cs:{14.70,-42}) -- cycle;
    \fill[blue!50,nearly transparent] (axis cs:{16.80,-42}) -- (axis cs:{16.80,20}) -- (axis 
    cs:{19.70,20}) -- (axis cs:{19.70,-42}) -- cycle;
    \legend{\footnotesize{$\mathbf{a}^{(1)}$}, \footnotesize{$\mathbf{a}^{(2)}$}, 
        \footnotesize{$\mathbf{a}^{(3)}$}};%
    \node [draw, fill=white] at (rel axis cs: 0.35,0.83) 
    {\shortstack[l]{\footnotesize{$\bar{a}$}}};
    \node [draw, fill=white] at (rel axis cs: 0.35,0.285) 
    {\shortstack[l]{\footnotesize{$\underline{a}$}}};
    \node at (rel axis cs: 0.28,0.95) 
    {\shortstack[l]{\footnotesize{\color{blue!50}{H2}}}};
    \node at (rel axis cs: 0.565,0.95) 
    {\shortstack[l]{\footnotesize{\color{green!100!black!100}{RS}}}};
    \node at (rel axis cs: 0.80,0.95) 
    {\shortstack[l]{\footnotesize{\color{blue!50}{H1}}}};
    \end{axis}		
    \end{tikzpicture}
    }
\end{subfigure}
\vspace{-1.5em}

%% file: main.bbl
\begin{thebibliography}{10}
\providecommand{\url}[1]{#1}
\csname url@samestyle\endcsname
\providecommand{\newblock}{\relax}
\providecommand{\bibinfo}[2]{#2}
\providecommand{\BIBentrySTDinterwordspacing}{\spaceskip=0pt\relax}
\providecommand{\BIBentryALTinterwordstretchfactor}{4}
\providecommand{\BIBentryALTinterwordspacing}{\spaceskip=\fontdimen2\font plus
\BIBentryALTinterwordstretchfactor\fontdimen3\font minus
  \fontdimen4\font\relax}
\providecommand{\BIBforeignlanguage}[2]{{%
\expandafter\ifx\csname l@#1\endcsname\relax
\typeout{** WARNING: IEEEtran.bst: No hyphenation pattern has been}%
\typeout{** loaded for the language `#1'. Using the pattern for}%
\typeout{** the default language instead.}%
\else
\language=\csname l@#1\endcsname
\fi
#2}}
\providecommand{\BIBdecl}{\relax}
\BIBdecl

\bibitem{OlleroTRO2022}
A.~Ollero \emph{et~al.}, ``{Past, Present, and Future of Aerial Robotic
  Manipulators},'' \emph{IEEE Transactions on Robotics}, vol.~38, no.~1, pp.
  626--645, 2022.

\bibitem{Alcantara2020IEEEAccess}
A.~Alcántara \emph{et~al.}, ``{Autonomous Execution of Cinematographic Shots
  With Multiple Drones},'' \emph{IEEE Access}, vol.~8, pp. 201\,300--201\,316,
  2020.

\bibitem{Tognon2019RAL}
M.~Tognon \emph{et~al.}, ``{A Truly-Redundant Aerial Manipulator System With
  Application to Push-and-Slide Inspection in Industrial Plants},'' \emph{IEEE
  Robotics and Automation Letters}, vol.~4, no.~2, pp. 1846--1851, 2019.

\bibitem{SilanoICUAS2021}
G.~Silano \emph{et~al.}, ``{A Multi-Layer Software Architecture for Aerial
  Cognitive Multi-Robot Systems in Power Line Inspection Tasks},'' in
  \emph{2021 International Conference on Unmanned Aircraft Systems}, 2021, pp.
  1624--1629.

\bibitem{BonillaICUAS23}
D.~{Bonilla Licea} \emph{et~al.}, ``{Communications-Aware Robotics: Challenges
  and Opportunities},'' in \emph{2023 International Conference on Unmanned
  Aircraft Systems}, 2023, {To Appear}.

\bibitem{Afifi2022ICRA}
A.~Afifi \emph{et~al.}, ``{Toward Physical Human-Robot Interaction Control with
  Aerial Manipulators: Compliance, Redundancy Resolution, and Input Limits},''
  in \emph{2022 International Conference on Robotics and Automation}, 2022, pp.
  4855--4861.

\bibitem{Corsini2022IROS}
G.~Corsini \emph{et~al.}, ``{Nonlinear Model Predictive Control for Human-Robot
  Handover with Application to the Aerial Case},'' in \emph{2022 IEEE
  International Conference on Intelligent Robots and Systems}, 2022, pp.
  7597--7604.

\bibitem{WojciechowskaHRI2019}
A.~Wojciechowska \emph{et~al.}, ``{Collocated Human-Drone Interaction:
  Methodology and Approach Strategy},'' in \emph{2019 14th IEEE International
  Conference on Human-Robot Interaction}, 2019, pp. 172--181.

\bibitem{Ajoudani2018AR}
A.~{Ajoudani} \emph{et~al.}, ``{Progress and prospects of the human-robot
  collaboration},'' \emph{Autonomous Robots}, vol.~42, no.~5, pp. 957--975,
  2018.

\bibitem{Gienger2018IROS}
M.~Gienger \emph{et~al.}, ``{Human-Robot Cooperative Object Manipulation with
  Contact Changes},'' in \emph{2018 IEEE International Conference on
  Intelligent Robots and Systems}, 2018, pp. 1354--1360.

\bibitem{Ortenzi2021TRO}
V.~Ortenzi \emph{et~al.}, ``{Object Handovers: A Review for Robotics},''
  \emph{IEEE Transactions on Robotics}, vol.~37, no.~6, pp. 1855--1873, 2021.

\bibitem{Belta2007RAM}
C.~Belta \emph{et~al.}, ``{Symbolic planning and control of robot motion [Grand
  Challenges of Robotics]},'' \emph{IEEE Robotics \& Automation Magazine},
  vol.~14, no.~1, pp. 61--70, 2007.

\bibitem{maler2004FTMATFTS}
O.~Maler \emph{et~al.}, ``{Monitoring temporal properties of continuous
  signals},'' in \emph{Formal Techniques, Modelling and Analysis of Timed and
  Fault-Tolerant Systems}.\hskip 1em plus 0.5em minus 0.4em\relax Berlin,
  Heidelberg: Springer, 2004, pp. 152--166.

\bibitem{Medina2016Humanoids}
J.~R. Medina \emph{et~al.}, ``{A human-inspired controller for fluid
  human-robot handovers},'' in \emph{2016 IEEE-RAS 16th International
  Conference on Humanoid Robots}, 2016, pp. 324--331.

\bibitem{Sisbot2012TRO}
E.~A. Sisbot \emph{et~al.}, ``{A Human-Aware Manipulation Planner},''
  \emph{IEEE Transactions on Robotics}, vol.~28, no.~5, pp. 1045--1057, 2012.

\bibitem{Peternel2017Humanoids}
L.~Peternel \emph{et~al.}, ``{Towards ergonomic control of human-robot
  co-manipulation and handover},'' in \emph{2017 IEEE 17th International
  Conference on Humanoid Robotics}, 2017, pp. 55--60.

\bibitem{Kshirsagar2019IROS}
A.~Kshirsagar \emph{et~al.}, ``{Specifying and Synthesizing Human-Robot
  Handovers},'' in \emph{2019 IEEE International Conference on Intelligent
  Robots and Systems}, 2019, pp. 5930--5936.

\bibitem{Webster2019ArXiv}
M.~{Webster} \emph{et~al.}, ``{An assurance-based approach to verification and
  validation of human–robot teams},'' \emph{arXiv preprint arXiv:1608.07403},
  September 2019.

\bibitem{Alpern1985IPL}
B.~Alpern \emph{et~al.}, ``{Defining liveness},'' \emph{Information Processing
  Letters}, vol.~21, pp. 181--185, 1985.

\bibitem{SilanoRAL2021}
G.~Silano \emph{et~al.}, ``{Power Line Inspection Tasks With Multi-Aerial Robot
  Systems Via Signal Temporal Logic Specifications},'' \emph{IEEE Robotics and
  Automation Letters}, vol.~6, no.~2, pp. 4169--4176, 2021.

\bibitem{CaballeroAR2022}
A.~{Caballero} \emph{et~al.}, ``{A Signal Temporal Logic Motion Planner for
  Bird Diverter Installation Tasks with Multi-Robot Aerial Systems},''
  \emph{ArXiv, 2210.09750}, October 2022.

\bibitem{MichielettoTRO2018}
G.~Michieletto \emph{et~al.}, ``{Fundamental Actuation Properties of
  Multirotors: Force–Moment Decoupling and Fail–Safe Robustness},''
  \emph{IEEE Transactions on Robotics}, vol.~34, no.~3, pp. 702--715, 2018.

\bibitem{LaValleBook}
S.~M. LaValle, \emph{{Sampling-Based Motion Planning}}.\hskip 1em plus 0.5em
  minus 0.4em\relax Cambridge University Press, 2006.

\bibitem{donze2010ICFMATS}
A.~Donz{\'e} \emph{et~al.}, ``{Robust satisfaction of temporal logic over
  real-valued signals},'' in \emph{{International Conference on Formal Modeling
  and Analysis of Timed Systems}}.\hskip 1em plus 0.5em minus 0.4em\relax
  Springer, 2010, pp. 92--106.

\bibitem{Fainekos2009TCS}
G.~E. {Fainekos} \emph{et~al.}, ``{Robustness of temporal logic specifications
  for continuous-time signals},'' \emph{Theoretical Computer Science}, vol.
  410, no.~42, pp. 4262--4291, 2009.

\bibitem{Pant2017CCTA}
Y.~V. Pant \emph{et~al.}, ``{Smooth operator: Control using the smooth
  robustness of temporal logic},'' in \emph{2017 IEEE Conference on Control
  Technology and Applications}, 2017, pp. 1235--1240.

\bibitem{MehdipourACC2019}
N.~Mehdipour \emph{et~al.}, ``{Arithmetic-Geometric Mean Robustness for Control
  from Signal Temporal Logic Specifications},'' in \emph{2019 American Control
  Conference}, 2019, pp. 1690--1695.

\bibitem{LindemannAutomatica2019}
L.~Lindemann \emph{et~al.}, ``{Robust control for signal temporal logic
  specifications using discrete average space robustness},'' \emph{Automatica},
  vol. 101, pp. 377--387, 2019.

\bibitem{Bertsekas2012Book}
D.~Bertsekas, \emph{{Dynamic programming and optimal control}}, Athena
  Scientific, 2012.

\bibitem{CaiMacroeconomicDynamics2017}
Y.~Cai \emph{et~al.}, ``{A Nonlinear Programming Method For Dynamic
  Programming},'' \emph{Macroeconomic Dynamics}, vol.~21, no.~2, pp. 336--361,
  2017.

\end{thebibliography}
